\documentclass{article}

\usepackage{arxiv}
\usepackage[utf8]{inputenc} 
\usepackage[T1]{fontenc}    
\usepackage{hyperref}       
\usepackage{url}            
\usepackage{booktabs}       
\usepackage{amsfonts}       
\usepackage{nicefrac}       
\usepackage{microtype}      
\usepackage{lipsum}		
\usepackage{graphicx}
\usepackage{natbib}
\usepackage{doi}
\usepackage{ulem}
\usepackage{comment}
\usepackage{xcolor}

\usepackage{booktabs} 
\usepackage{tabularx} 
\usepackage{multirow}
\usepackage{caption}
\usepackage{listings}
\usepackage{xcolor}
\usepackage{algorithm}
\usepackage{algpseudocode}
\usepackage{caption}
\usepackage{inconsolata} 

\usepackage{amsmath}

\usepackage{pifont}
\newcommand{\cmark}{\ding{51}} 
\newcommand{\xmark}{\ding{55}} 

\newcommand{\revs}[1]{\textcolor{black}{#1}}

\lstdefinestyle{pythonstyle}{
    language=Python,
    basicstyle=\ttfamily\small,
    keywordstyle=\color{blue},
    commentstyle=\color{gray},
    stringstyle=\color{teal!70!black},
    showstringspaces=false,
    breaklines=true,
    frame=single,
    framerule=0.5pt,
    rulecolor=\color{black!40},
    tabsize=4,
    belowskip=1.5em,
    aboveskip=1.0em
}

\lstdefinestyle{promptstyle}{
    language={},
    basicstyle=\ttfamily\small,
    breaklines=true,
    breakatwhitespace=true,
    columns=fullflexible,
    frame=single,
    framerule=0.5pt,
    rulecolor=\color{black!40},
    backgroundcolor=\color{gray!5},
    tabsize=4,
    showstringspaces=false,
    belowskip=1.5em,
    aboveskip=1.0em,
    literate={-}{-}1
             {\\theta}{{$\theta$}}1
             {\\lambda}{{$\lambda$}}1
             {\\Gamma}{{$\Gamma$}}1
             {\\times}{{$\times$}}1
             {\\ge}{{$\ge$}}1
             {\\le}{{$\le$}}1
             {\\nu}{{$\nu$}}1
             {\\xi}{{$\xi$}}1
             {\\eta}{{$\eta$}}1
             {\\pm}{{$\pm$}}1
             {\\to}{{$\to$}}1
             {\\partial}{{$\partial$}}1
             {\\Sigma}{{$\Sigma$}}1
             {\\sqrt}{{$\sqrt{}$}}1  
}

\usepackage[most]{tcolorbox}
\usepackage{xcolor}
\usepackage{enumitem}

\definecolor{promptbg}{RGB}{233,240,255}
\definecolor{annotbg}{RGB}{255,250,200}
\definecolor{highlight}{RGB}{252,213,122}

\title{FEM-Bench: A Structured Scientific Reasoning Benchmark for Evaluating Code-Generating LLMs}

\author{ \href{https://orcid.org/0000-0001-9879-044X}{\includegraphics[scale=0.06]{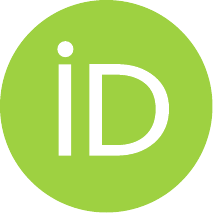}\hspace{1mm}Saeed Mohammadzadeh}\thanks{Saeed Mohammadzadeh and Erfan Hamdi are co-first authors} \\
	Department of Mechanical Engineering\\
	Boston University\\
	Boston, MA \\
	\texttt{saeedmhz@bu.edu} \\
	\And
	\href{https://orcid.org/0009-0008-3262-7533?}{\includegraphics[scale=0.06]{orcid.pdf}\hspace{1mm}Erfan Hamdi$^*$} \\
	Department of Mechanical Engineering\\
	Boston University\\
	Boston, MA \\
	\texttt{erfan@bu.edu} \\
    \And
    \href{ORCID?}{\includegraphics[scale=0.06]{orcid.pdf}\hspace{1mm}Joel Shor} \\
	1. Move37 Labs\\
	2. Department of Mechanical Engineering, Boston Univeristy\\
	\texttt{joel.shor@move37labs.ai} \\
    \And
    \href{https://orcid.org/0000-0001-8099-3468}{\includegraphics[scale=0.06]{orcid.pdf}\hspace{1mm}Emma Lejeune} \\
	Department of Mechanical Engineering\\
	Boston University\\
	Boston, MA \\
	\texttt{elejeune@bu.edu} \\
}

\renewcommand{\shorttitle}{FEM-Bench 2025}

\hypersetup{
pdftitle={A template for the arxiv style},
pdfsubject={q-bio.NC, q-bio.QM},
pdfauthor={David S.~Hippocampus, Elias D.~Striatum},
pdfkeywords={First keyword, Second keyword, More},
}

\begin{document}
\maketitle

\begin{abstract}
As LLMs advance their reasoning capabilities about the physical world, the absence of rigorous benchmarks for evaluating their ability to generate scientifically valid physical models has become a critical gap. Computational mechanics, the discipline that develops and applies mathematical models and numerical methods to predict the behavior of physical systems under forces, deformation, and constraints, provides an ideal foundation for structured scientific reasoning based evaluation. Problems follow clear mathematical structure, enforce strict physical and numerical constraints, and support objective verification. The discipline also requires constructing explicit models of physical systems and reasoning about geometry, spatial relationships, and material behavior, which connects directly to emerging goals in AI related to physical reasoning and world modeling. We introduce FEM-Bench, a computational mechanics benchmark designed to evaluate the ability of LLMs to generate correct finite element method (FEM) and related code. FEM-Bench 2025 contains a suite of introductory but nontrivial tasks aligned with material from a first graduate course on computational mechanics. These tasks capture essential numerical and physical modeling challenges while representing only a small fraction of the complexity present in the discipline. Despite their simplicity, state-of-the-art LLMs do not reliably solve all of them. In a five attempt run, the best performing model at function writing, Gemini 3 Pro, completed 30/33 tasks at least one out of five times, and 26/33 tasks five out of five times. The best performing model at unit test writing, GPT-5, had an Average Joint Success Rate of 73.8\%.  Other popular models showed a broad range of performance on the benchmark. FEM-Bench establishes a structured foundation for evaluating AI-generated scientific code, and future iterations will incorporate increasingly sophisticated tasks to track progress as models evolve.
\end{abstract}


\section{Introduction}
\label{sec:intro}
Modern AI systems are increasingly evaluated on their ability to build internal models of the physical world, including reasoning about forces, geometry, spatial relationships, and the mechanical behavior of physical systems \citep{bakhtin2019phyre, wang2023newton}. Simultaneously, large language models (LLMs) show growing potential to assist with scientific and engineering workflows, including reasoning about physical systems and automated generation of simulation and analysis code \citep{cui2025curie}. As these models advance, a central question emerges: can LLMs produce implementations of physics-based numerical methods that are correct, reliable, and scientifically meaningful? Existing code-generation benchmarks test general programming logic \citep{austin2021program, chen2021codex}, software engineering skills \citep{jimenez2023swe}, and mathematical reasoning \citep{glazer2024frontiermath}, but no widely used benchmarks evaluate an LLM’s ability to carry out the physical modeling, numerical discretization, and structured computational reasoning required for advanced scientific computing \citep{tian2024scicode}. Computational mechanics, the discipline that formulates and solves mathematical models of physical systems using numerical methods, provides a natural and rigorous domain in which to investigate these capabilities.

Computational-mechanics--based tasks represent a particularly important frontier for LLM capabilities \citep{jiang2025deepseek}. Physics-based simulations grounded in this discipline underpin real-world applications in domains as diverse as robotics \citep{huang2020dynamic}, digital twins of aircraft and human systems \citep{niederer2021scaling}, climate modeling \citep{danabasoglu2020community}, and engineering design and optimization \citep{talischi2012polytop}. Across these areas, several themes are consistent: successful simulation requires the seamless integration of mathematics, physics, geometry, numerical methods, and programming, and correctness is non-negotiable. A program that runs but produces a non-convergent, unstable, or physically impossible result is of no scientific value \citep{peng2011reproducible}. For this reason, computational mechanics has a long and mature tradition of verification, validation, and uncertainty quantification \citep{oberkampf2010verification}, providing structured, quantitative tools for assessing whether a model is numerically consistent, physically plausible, and predictive. These practices make the domain especially well-suited for evaluating LLM-generated code, because failures are interpretable, quantitatively measurable, and can be traced not only to programming or structural errors but also to deeper mistakes in implementing governing equations, enforcing numerical consistency, handling geometric transformations, or managing floating-point--sensitive operations such as numerical integration and matrix assembly \citep{arndt2023deal, alnaes2015fenics}.

Thus, this branch of scientific computing provides a well-structured and scientifically grounded testbed for critically evaluating the emerging capabilities of LLMs. Problems in computational mechanics follow a precise mathematical pipeline: e.g., governing equations are formulated, converted into weak or variational forms, discretized into finite-dimensional approximations, and assembled into global algebraic systems \citep{courant1994variational,turner1956stiffness}. Solutions are then validated through objective numerical checks such as symmetry, consistency, and mesh convergence \citep{zienkiewicz1997finite}. From the perspective of LLM reasoning, these tasks span multiple forms of structured cognition: hierarchical reasoning in which local element routines combine into global models; geometric transformations across coordinate systems; the construction of explicit models of physical systems; and careful handling of numerical integration, stability, and floating-point--sensitive operations. Moreover, the computational mechanics literature is extensive and mature. The computational mechanics literature spans classic textbooks \citep{hughes2003finite}, widely taught graduate curricula in engineering \citep{garikipati_fem_coursera, shojaei2025ai}, and ongoing research at the frontier of numerical methods development \citep{kamarei2026nine}, offering a rich, well-understood space of problems with clear expectations for correctness and implementation quality \citep{szabo2021finite}.

In this work, we introduce FEM-Bench, a benchmark designed to evaluate the ability of LLMs to generate correct implementations of finite element method (FEM) and related computational mechanics tasks. 
\revs{FEM-Bench is designed as a diagnostic challenge suite rather than a large-scale training dataset, prioritizing interpretability and reasoning depth over task count to enable fine-grained analysis of LLM failure modes in structured scientific computing.}
FEM-Bench 2025 focuses on introductory but nontrivial tasks aligned with material from a first graduate course on FEM. 
These tasks isolate essential numerical and physical modeling challenges while remaining amenable to automated evaluation. Despite their relative simplicity, we find that state-of-the-art LLMs show significant room for improvement, revealing significant gaps between current model capabilities and the requirements of physics-based scientific computing. 

The contributions of this paper are threefold. First, we present the FEM-Bench framework, including its design principles and scope. Second, we provide a suite of tasks and corresponding evaluation tools grounded in canonical computational mechanics concepts. Third, we conduct a baseline evaluation across several leading LLMs to characterize current performance. Looking forward, FEM-Bench establishes a foundation for increasingly challenging computational mechanics tasks in future iterations, enabling systematic tracking of progress in scientific computing specific code generation.

\section{Background}
\label{sec:background}
This Section provides background on computational mechanics as a domain for evaluating LLMs, emphasizing physical grounding, algorithmic structure, and rigorous verification culture. We then motivate the need for FEM-Bench as a benchmark designed to probe these capabilities in a scientifically meaningful setting.

\subsection{Computational Mechanics as a Test of Physical World Modeling}

Computational mechanics provides the mathematical and numerical foundation for simulating the behavior of physical systems under forces, deformation, and constraints \citep{belytschko2014nonlinear, gurtin2010mechanics}. At its core, the field translates physical laws into solvable mathematical models, which are then discretized and implemented as algorithms suitable for computation \citep{langtangen2003computational}. In doing so, computational mechanics serves as a direct test of whether a model can correctly represent and reason about the physical world in algorithmic form. Among the many techniques used in computational mechanics, the Finite Element Method (FEM) and Matrix Structural Analysis (MSA) are two of the most widely taught and broadly applied approaches \citep{hughes2003finite,mcguire2000matrix}. MSA predates FEM and was originally developed to analyze framed structures such as trusses and beams using stiffness matrices derived directly from structural mechanics \citep{argyris1960energy,turner1956stiffness}. Although more specialized in scope, MSA shares the same foundational principles as FEM--local element stiffness relations, coordinate transformations, and global assembly--and is often viewed as an early, specialized precursor to the broader finite element framework \citep{cook2007concepts}. These shared ideas make MSA a natural companion to FEM in introductory mechanics curricula and a relevant component of the FEM-Bench task space \citep{felippa2004introduction}.
\begin{figure}[ht]
        \centering
        \includegraphics[width=0.95\linewidth]{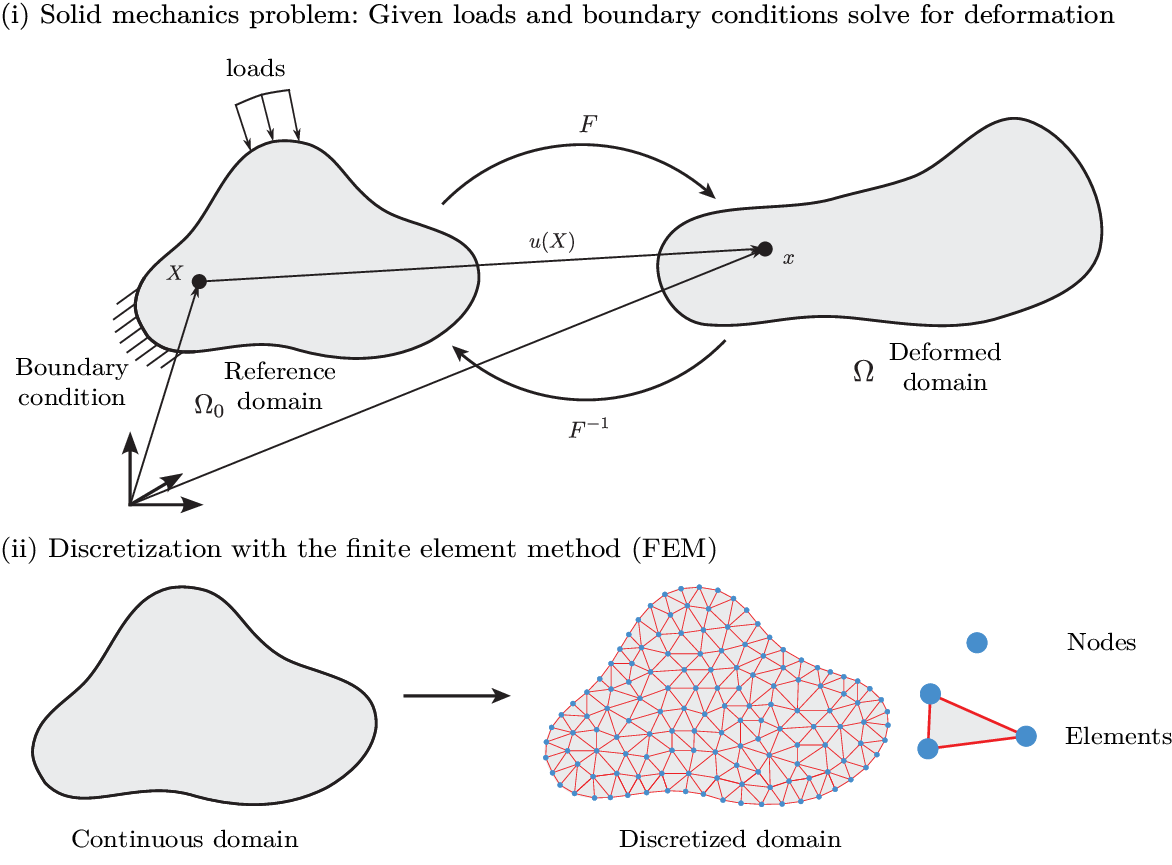}
        \caption{Schematic overview of the finite element method (FEM). 
        (i) A solid mechanics problem: given loads and boundary conditions on the 
        reference configuration, solve for the deformation mapping to the deformed 
        configuration. (ii) Discretization step: the continuous domain is 
        approximated by a mesh of finite elements and nodes, enabling numerical 
        solution of the relevant governing equations.}
        \label{fig:fem_schematic}
\end{figure}
\subsection{Computational Mechanics as a Test of Structured, Multistep Reasoning}

At a high level, FEM proceeds through a well-defined sequence of modeling steps. Each step depends on the correct execution of preceding stages, requiring careful composition of intermediate representations and computations. As illustrated in Fig. \ref{fig:fem_schematic}, a continuous physical domain is first subdivided into a finite mesh of elements. Physical governing equations are expressed in a weak or variational form, enabling approximate solutions in finite-dimensional function spaces \citep{strang1973analysis}. On each element, basis (or “shape”) functions define how the unknown field varies locally, and numerical quadrature is used to evaluate integrals appearing in the weak form \citep{hughes2003finite,press2007numerical}. These element-level contributions are then assembled into a global algebraic system whose solution approximates the physical equilibrium state or transient evolution of the system \citep{bathe1996finite}. A schematic illustration of the discretization appears in Fig. \ref{fig:fem_schematic}. This mathematical pipeline is central to nearly all engineering simulation software and forms the conceptual backbone of FEM-Bench.

The mechanics tasks underlying FEM, and by extension MSA, involve several forms of structured computational reasoning that are highly relevant for evaluating modern LLMs \citep{tian2024scicode}. Hierarchical reasoning is required because global behavior emerges from local element routines combined through assembly operators \citep{arndt2023deal}. Geometric reasoning appears through coordinate transformations, Jacobians, and mappings between reference and global coordinate systems \citep{cottrell2009isogeometric}. Numerical reasoning arises through quadrature rules, element stiffness derivations, matrix assembly, stability considerations, and floating-point--sensitive calculations \citep{higham2002accuracy}. Finally, physical reasoning is inherent to the discipline: implementations must faithfully encode conservation laws, constitutive relations, and boundary conditions \citep{holzapfel2002nonlinear}. These components appear even in introductory FEM tasks, making the domain particularly suitable for probing LLM capabilities in scientific computing.

\subsection{Verification and Evaluation in Computational Mechanics}

An essential pillar of computational mechanics, equally relevant for benchmarking, is its strong culture of verification, validation, and uncertainty quantification (VVUQ) \citep{oberkampf2010verification}. These practices transform numerical implementations into testable artifacts with well-defined correctness criteria, making them particularly suitable for diagnostic evaluation. Because simulation outputs must reflect real physical behavior, the field has developed rigorous practices for assessing correctness, including patch tests, equilibrium checks, symmetry requirements, energy consistency, and mesh convergence studies \citep{belytschko2014nonlinear, cook2007concepts, macneal1985proposed,roache1998verification}. These methods provide objective, quantitative signals of whether an implementation is functioning correctly. For evaluating LLM-generated code, such tests are especially valuable: they make failures interpretable and help distinguish mistakes in programming logic \citep{ammann2017introduction}, geometry handling \citep{foley1996computer}, numerical implementation \citep{trefethen2022numerical}, or floating-point--sensitive operations such as numerical integration and matrix assembly \citep{higham2002accuracy}.

\subsection{Motivation for FEM-Bench}

The broader landscape of existing LLM benchmarks highlights the need for a physics-based alternative. Popular code-generation benchmarks such as HumanEval~\citep{humaneval}, MBPP~\citep{mbpp}, SWE-Bench~\citep{jimenez2024swebench}, and DS-1000~\citep{ds1000} evaluate programming logic, tool use, or general software engineering skills. Datasets targeting physical reasoning often focus on qualitative judgments or simplified environments rather than implementation of scientific algorithms~\citep{hamdi2026towards,lejeune2020mechanical,nucleobench}. Likewise, benchmarks for mathematics~\citep{hendrycksmath2021} or symbolic reasoning~\citep{symbolic_reasoning} do not require translating equations into stable, reliable numerical code.
\revs{The application of fine-tuning open-weight models have also been explored in multiple recent works in the field. For example Deotale et al.~\cite{deotale2026all} experimented with fine-tuning several open-weight models on a dataset of hand crafted and LLM-generated finite element codes using FEniCS, finding that fine-tuning improved model performance. Similarly, Shojaei et al.~\cite{shojaei2025ai} successfully fine-tuned models to develop a course-specific ``expert model'' integrated into the AI-University educational platform.}
While recent efforts like FEABench~\citep{mudur2024feabench} introduce agent-based tasks that interact with professional simulation software, they primarily measure the ability to navigate complex software APIs and external tools; this highlights a critical need to decouple the interpretation of software documentation from the actual implementation of the underlying physical and numerical reasoning.
As a result, there is a limited availability of benchmarks to evaluate whether LLMs can generate the kinds of physics-grounded, numerically consistent programs that underpin modern scientific computing~\citep{tian2024scicode}.

Computational mechanics therefore provides both the conceptual structure and the evaluative tools needed for such an assessment~\citep{oberkampf2010verification}. Its combination of mathematical rigor, geometric complexity, numerical precision, and physical grounding makes it an ideal basis for the development of FEM-Bench~\citep{hughes2003finite,mcguire2000matrix}. 
\revs{FEM-Bench is designed to probe four LLM capabilities that rarely co-occur in existing benchmarks and that we revisit in detail in our error analysis (Section~\ref{sec:error}):}
\begin{itemize}
    \item \revs{\textbf{Domain knowledge:} the ability to recall sufficiently detailed knowledge of the underlying mechanics and numerical structures required by a task.Evaluated by tasks in which the reference implementation must be reproduced without scaffolding, e.g., Tier~3 (T3) variants that withhold helper functions, and standalone tasks such as \texttt{MSA\_3D\_local\_geometric\_stiffness}.}
    
    \item \revs{\textbf{Compositional reasoning:} the ability to combine and manipulate multiple components into a correct multi-step computation. Evaluated through the tiered helper-function design (T1/T2/T3) and through 
    tasks such as \texttt{MSA\_3D\_elastic\_critical\_load}, whose reference implementation chains together multiple physical steps (Fig.~\ref{fig:schematic}).}
    
    \item \revs{\textbf{Algorithmic fidelity:} the ability to implement a computation with the consistency required for it to execute and produce numerically correct outputs (correct indexing, consistent sign conventions, 
    complete routines, etc.). Evaluated by reference-output matching on curated verification inputs (Section~\ref{sec:function_correctness}).}
    
    \item \revs{\textbf{Self-verification:} the ability to express the correctness criteria of a task as discriminative, physics-aware unit tests 
    that both pass on a correct implementation and fail on known-incorrect implementations. Evaluated by the joint test success rate over expected-failure cases (Section~\ref{sec:test_suite_evaluation}).}
\end{itemize}
In the next Section, we introduce the benchmark design and describe the task suite that forms the FEM-Bench 2025 release.

\section{Methods}
\label{sec:methods}

\begin{figure}[h]
    \centering
    \includegraphics[width=0.65\linewidth]{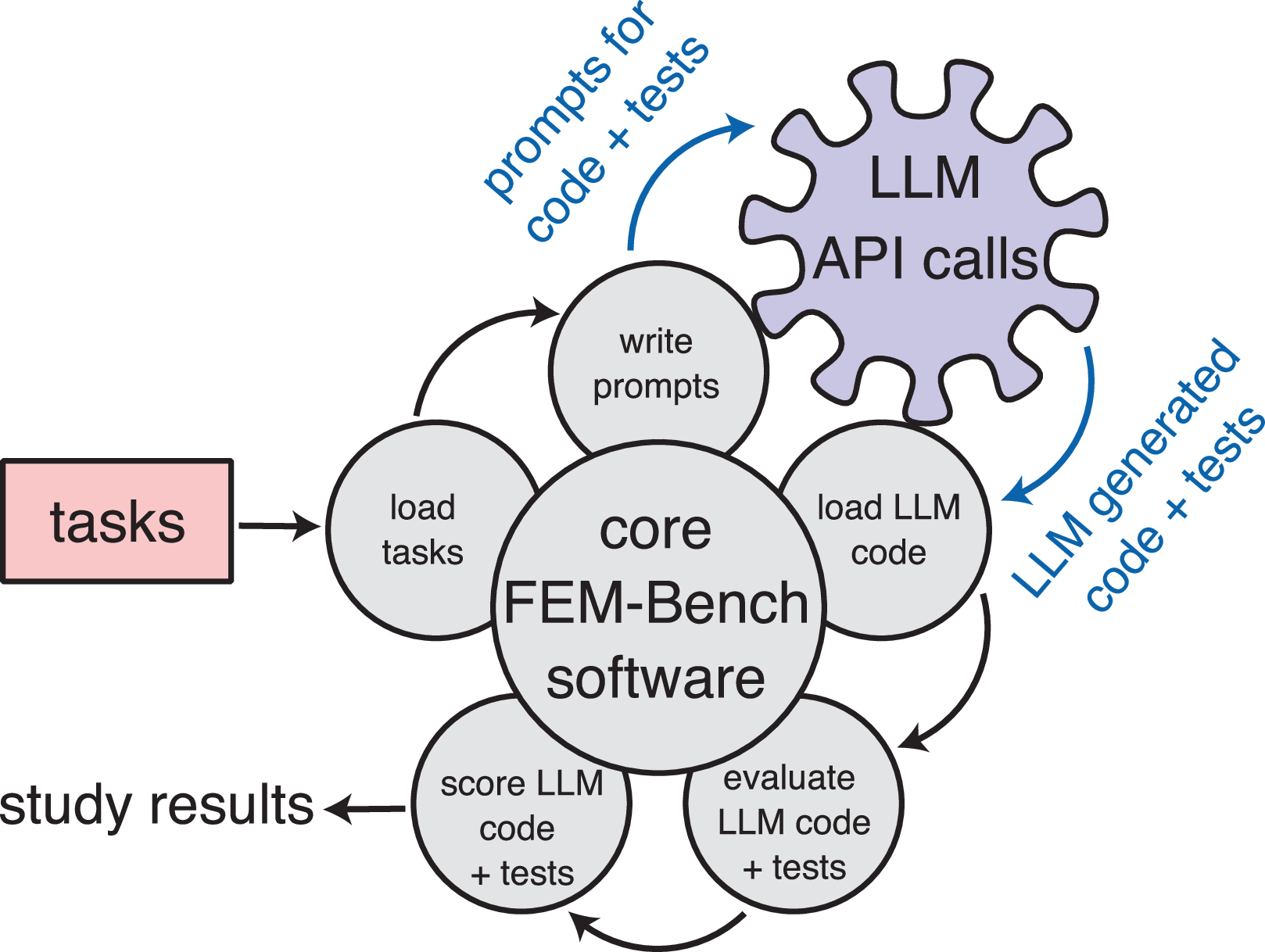}
    \caption{
        \textbf{Overview of the FEM-Bench workflow.}
        Tasks are defined as self-contained Python modules specifying reference implementations,
        dependencies, and unit tests. The core FEM-Bench software loads these tasks, constructs
        standardized prompts for both code-generation and test-generation, and interfaces with
       LLMs through model-specific API clients. LLM outputs (generated
        code and test suites) are then parsed, validated, and evaluated using reference-based
        numerical checks and expected-failure unit tests. The framework produces detailed scores
        for each model and task, enabling reproducible and interpretable comparison of LLM
        performance.}
    \label{fig:fembench_pipeline}
\end{figure}

This Section describes the FEM-Bench methodology, including the structure of the benchmark, the construction of tasks and prompts, the evaluation pipeline for code and test generation, and the initial selection of LLMs to evaluate. FEM-Bench follows a modular design: tasks are defined as self-contained Python modules, prompts are generated automatically from task metadata, LLM outputs are parsed and validated, and results are computed through reference-based numerical and unit-testing procedures. The framework is designed for reproducibility, extensibility, and principled evaluation. A high-level overview of the FEM-Bench workflow, including task loading, prompt generation, model inference, and evaluation, is shown in Fig.~\ref{fig:fembench_pipeline},
\revs{with the corresponding pseudocode provided in Algorithm \ref{alg:fembench_pipeline}.}

\algrenewcommand\algorithmicrequire{\textbf{Input:}}
\algrenewcommand\algorithmicensure{\textbf{Output:}}
\begin{algorithm}[H]
\caption{\revs{FEM-Bench Evaluation Pipeline}}
\label{alg:fembench_pipeline}
\begin{algorithmic}[1]
\Require Tasks $\mathcal{T}$, models $\mathcal{M}$
\Ensure Function Correctness score, Average joint test success rate
\State Load tasks
\State Generate task and test prompts
\ForAll{model $m \in \mathcal{M}$}
    \ForAll{task $t \in \mathcal{T}$}
        \State Generate code with $m$ using task prompt of $t$
        \State Generate tests with $m$ using test prompt of $t$
    \EndFor
\EndFor
\State Evaluate generated code against reference outputs
\State Evaluate generated tests on reference function and expected-failure implementations
\State Aggregate and save scores
\end{algorithmic}
\end{algorithm}

\subsection{Problem Definition}
\label{sec:problem_definition}

FEM-Bench evaluates two complementary capabilities of LLMs: the ability to generate numerically correct scientific code, and the ability to generate unit tests that verify that code. In this benchmark, producing correct code means generating a Python function that conforms to a prescribed signature, runs without syntax errors, and returns outputs that satisfy the mathematical and physical requirements of the task. Similarly, producing a correct unit test means generating tests written using \texttt{pytest}, a widely used Python testing framework, that execute without errors, pass on the reference implementation, and fail on a curated set of known incorrect implementations.
Because scientific software correctness depends as much on verification as on implementation, FEM-Bench treats code synthesis and test synthesis as intertwined, first-class components of the benchmark. Together, these two components define the core problem that FEM-Bench poses to LLMs: not simply producing code that runs, but producing code and unit tests that are technically correct.

\subsection{Task Definition and Structure}
\label{sec:task_definition}

FEM-Bench defines each task as a self-contained Python module with a standardized, explicit structure designed for transparency, reproducibility, and extensibility. 
\revs{With regard to scientific content, each task centers around a single Python function with a prescribed signature that computes a well-defined numerical or physical quantity. For example, the core content of a task may involve computing a local element stiffness matrix, evaluating shape functions and their derivatives, assembling a global stiffness matrix, or running a full analysis from start to finish. Within a given task, the goal will be to implement and test the prescribed core content. For core content where the typical implementation would involve many helper functions, such as elastic critical load analysis, which requires a linear elastic solve, geometric stiffness assembly, and a generalized eigenvalue solve, we often define separate tasks corresponding to different tiers of provided helper functions. This tiering system is the primary mechanism by which FEM-Bench modulates task difficulty and isolates specific reasoning demands. A ``higher-level'' task such as \texttt{elastic\_critical\_load} can then be evaluated with varying amounts of scaffolding, exposing whether a model's failure stems from either a specific sub-component, or compositional reasoning across components.}
\revs{
Unit tests, also contained within each task, are designed to probe correctness at multiple levels, including structural properties, consistency with sub-components, and agreement with analytical solutions. A task is considered complete when its unit tests collectively cover the most meaningful physical and numerical failure modes, as illustrated by the expected-failure implementations bundled with each task.}

Every task includes: a reference implementation with a detailed docstring (which becomes part of the LLM prompt) 
\revs{
specifying mathematical and physical requirements,} 
Pytest-style test functions with descriptive docstrings (also incorporated directly into the prompt), optional dependency functions, known incorrect implementations used as expected failures, and a single \texttt{task\_info()} function. The \texttt{task\_info()} function assembles these components into a structured dictionary that contains all metadata and source code required for prompt generation, execution, and evaluation. This ``source-first'' design ensures that tasks remain fully executable and easy to inspect or extend. The full task template is shown in Listing~\ref{lst:task_template}.

\revs{
All tasks in FEM-Bench 2025 are manually written by the authors, with LLM assistance limited to narrow sub-function contributions that were reviewed and verified. Each task begins from a working codebase developed via test-driven development and validated against multiple analytical solutions.
}
\begin{lstlisting}[style=pythonstyle, caption={FEM-Bench task template}, label={lst:task_template}]
import numpy as np

# === Dependency functions (if any) ===
def helper_1(...):
    ...
def helper_2(...):
    ...

# === Reference implementation ===
def main_fcn(...):
    """Compute or solve something."""
    ...

# === Test functions ===
def test_case_1(fcn):
    """Docstring explaining what is tested."""
    ...

# === Known failing examples (optional) ===
def fail_case_1(...):
    ...
def fail_case_2(...):
    ...

# === task_info() metadata ===
def task_info():
    task_id = "unique_task_name"
    task_short_description = "concise description of what the task does"
    created_date = "YYYY-MM-DD"
    created_by = "your_name"

    main_fcn = main_fcn
    required_imports = [
        "import numpy as np",
        "import pytest",
        # additional imports if needed
    ]

    fcn_dependencies = [helper_1, helper_2]  # or [] if none

    reference_verification_inputs = [
        # List of lists: each sublist contains args for main_fcn
        [arg1, arg2, ...],
        ...
    ]

    test_cases = [
        {
            "test_code": test_case_1,
            "expected_failures": [fail_case_1, fail_case_2]  # or []
        },
        ...
    ]

    return {
        "task_id": task_id,
        "task_short_description": task_short_description,
        "created_date": created_date,
        "created_by": created_by,
        "main_fcn": main_fcn,
        "required_imports": required_imports,
        "fcn_dependencies": fcn_dependencies,
        "reference_verification_inputs": reference_verification_inputs,
        "test_cases": test_cases,
        "allow_negation_for_match": False,
        "python_version": "version_number",
        "package_versions": {"numpy": "version"},
    }
\end{lstlisting}

When a task is loaded, FEM-Bench extracts and normalizes the source code for the reference function, helpers, and tests, and stores them in a \texttt{Task} object defined in \texttt{task\_base.py}. This object provides the canonical representation of the task throughout the remainder of the pipeline, including prompt construction, LLM inference, and evaluation.

\subsubsection{Example Task}

To ground this template in a concrete example, Listing~\ref{lst:example_task_msa_3d} shows an actual FEM-Bench task from the FEM-Bench 2025 suite. This task defines the local $12\times 12$ elastic stiffness matrix for a 3D Euler--Bernoulli beam element, along with two pytest-style test functions. The first test (\texttt{test\_local\_stiffness\_3D\_beam}) checks structural properties such as symmetry, rigidity, and consistency of axial, torsional, and bending terms. The second test (\texttt{test\_cantilever\_deflection\_matches\_euler\_bernoulli}) compares numerical displacements with closed-form Euler--Bernoulli beam theory under different loading directions. The \texttt{task\_info()} function then packages the reference implementation, verification inputs, and expected-failure implementations into the standardized FEM-Bench task format.
\revs{Supplementary examples of test cases and expected failure functions are shown in Appendix \ref{apx:tests_expected_fail}, in Table \ref{tab:test_details} and \ref{tab:expected_failures} respectively. All complete task implementations used in this study are published on the FEM-Bench GitHub.}

\begin{lstlisting}[style=pythonstyle,
    caption={Example FEM-Bench task: local stiffness matrix for a 3D Euler--Bernoulli beam element.},
    label={lst:example_task_msa_3d}
]
import numpy as np


def MSA_3D_local_elastic_stiffness_CC0_H0_T0(
    E: float,
    nu: float,
    A: float,
    L: float,
    Iy: float,
    Iz: float,
    J: float
) -> np.ndarray:
    """
    Return the 12x12 local elastic stiffness matrix for a 3D Euler-Bernoulli beam element.

    The beam is assumed to be aligned with the local x-axis. The stiffness matrix
    relates local nodal displacements and rotations to forces and moments using the equation:

        [force_vector] = [stiffness_matrix] @ [displacement_vector]

    Degrees of freedom are ordered as:
        [u1, v1, w1, θx1, θy1, θz1, u2, v2, w2, θx2, θy2, θz2]

    Where:
        - u, v, w: displacements along local x, y, z
        - θx, θy, θz: rotations about local x, y, z
        - Subscripts 1 and 2 refer to node i and node j of the element

    Parameters:
        E (float): Young's modulus
        nu (float): Poisson's ratio (used for torsion only)
        A (float): Cross-sectional area
        L (float): Length of the beam element
        Iy (float): Second moment of area about the local y-axis
        Iz (float): Second moment of area about the local z-axis
        J (float): Torsional constant

    Returns:
        np.ndarray: A 12x12 symmetric stiffness matrix representing axial, torsional,
                    and bending stiffness in local coordinates.
    """
    k_e = np.zeros((12, 12))
    # Axial terms - extension of local x axis
    axial_stiffness = E * A / L
    k_e[0, 0] = axial_stiffness
    k_e[0, 6] = -axial_stiffness
    k_e[6, 0] = -axial_stiffness
    k_e[6, 6] = axial_stiffness
    # Torsion terms - rotation about local x axis
    torsional_stiffness = E * J / (2.0 * (1 + nu) * L)
    k_e[3, 3] = torsional_stiffness
    k_e[3, 9] = -torsional_stiffness
    k_e[9, 3] = -torsional_stiffness
    k_e[9, 9] = torsional_stiffness
    # Bending terms - bending about local z axis
    k_e[1, 1] = E * 12.0 * Iz / L ** 3.0
    k_e[1, 7] = E * -12.0 * Iz / L ** 3.0
    k_e[7, 1] = E * -12.0 * Iz / L ** 3.0
    k_e[7, 7] = E * 12.0 * Iz / L ** 3.0
    k_e[1, 5] = E * 6.0 * Iz / L ** 2.0
    k_e[5, 1] = E * 6.0 * Iz / L ** 2.0
    k_e[1, 11] = E * 6.0 * Iz / L ** 2.0
    k_e[11, 1] = E * 6.0 * Iz / L ** 2.0
    k_e[5, 7] = E * -6.0 * Iz / L ** 2.0
    k_e[7, 5] = E * -6.0 * Iz / L ** 2.0
    k_e[7, 11] = E * -6.0 * Iz / L ** 2.0
    k_e[11, 7] = E * -6.0 * Iz / L ** 2.0
    k_e[5, 5] = E * 4.0 * Iz / L
    k_e[11, 11] = E * 4.0 * Iz / L
    k_e[5, 11] = E * 2.0 * Iz / L
    k_e[11, 5] = E * 2.0 * Iz / L
    # Bending terms - bending about local y axis
    k_e[2, 2] = E * 12.0 * Iy / L ** 3.0
    k_e[2, 8] = E * -12.0 * Iy / L ** 3.0
    k_e[8, 2] = E * -12.0 * Iy / L ** 3.0
    k_e[8, 8] = E * 12.0 * Iy / L ** 3.0
    k_e[2, 4] = E * -6.0 * Iy / L ** 2.0
    k_e[4, 2] = E * -6.0 * Iy / L ** 2.0
    k_e[2, 10] = E * -6.0 * Iy / L ** 2.0
    k_e[10, 2] = E * -6.0 * Iy / L ** 2.0
    k_e[4, 8] = E * 6.0 * Iy / L ** 2.0
    k_e[8, 4] = E * 6.0 * Iy / L ** 2.0
    k_e[8, 10] = E * 6.0 * Iy / L ** 2.0
    k_e[10, 8] = E * 6.0 * Iy / L ** 2.0
    k_e[4, 4] = E * 4.0 * Iy / L
    k_e[10, 10] = E * 4.0 * Iy / L
    k_e[4, 10] = E * 2.0 * Iy / L
    k_e[10, 4] = E * 2.0 * Iy / L
    return k_e


def test_local_stiffness_3D_beam(fcn):
    """
    Comprehensive test for local_elastic_stiffness_matrix_3D_beam:
    - shape check
    - symmetry
    - expected singularity due to rigid body modes
    - block-level verification of axial, torsion, and bending terms
    """
    # Beam properties
    E = 200e9         # Young's modulus
    nu = 0.3          # Poisson's ratio
    A = 0.01          # Cross-sectional area
    L = 2.0           # Length of the beam
    Iy = 8         # Moment of inertia about y
    Iz = 6         # Moment of inertia about z
    J = 1          # Torsional constant

    k = fcn(E, nu, A, L, Iy, Iz, J)

    # --- Shape check ---
    assert k.shape == (12, 12)

    # --- Symmetry check ---
    assert np.allclose(k, k.T, atol=1e-12)

    # --- Singularity check (due to 6 rigid-body modes) ---
    eigvals = np.linalg.eigvalsh(k)
    min_eigval = np.min(np.abs(eigvals))
    assert min_eigval < 1e-10, f"Expected a zero eigenvalue, but smallest was {min_eigval:.2e}"

    # --- Axial terms block ---
    expected_axial = E * A / L
    assert np.isclose(k[0, 0], expected_axial, rtol=1e-12)
    assert np.isclose(k[0, 6], -expected_axial, rtol=1e-12)
    assert np.isclose(k[6, 0], -expected_axial, rtol=1e-12)
    assert np.isclose(k[6, 6], expected_axial, rtol=1e-12)

    # --- Torsional terms block (theta_x DOFs) ---
    G = E / (2 * (1 + nu))
    expected_torsion = G * J / L
    assert np.isclose(k[3, 3], expected_torsion, rtol=1e-12)
    assert np.isclose(k[3, 9], -expected_torsion, rtol=1e-12)
    assert np.isclose(k[9, 3], -expected_torsion, rtol=1e-12)
    assert np.isclose(k[9, 9], expected_torsion, rtol=1e-12)

    # --- Bending about local z (v--theta_z: DOFs 1, 5, 7, 11) ---
    expected_bz_11 = E * 12.0 * Iz / L**3
    expected_bz_15 = E * 6.0 * Iz / L**2
    expected_bz_55 = E * 4.0 * Iz / L
    expected_bz_511 = E * 2.0 * Iz / L
    assert np.isclose(k[1, 1], expected_bz_11, rtol=1e-12)
    assert np.isclose(k[1, 5], expected_bz_15, rtol=1e-12)
    assert np.isclose(k[5, 5], expected_bz_55, rtol=1e-12)
    assert np.isclose(k[5, 11], expected_bz_511, rtol=1e-12)

    # --- Bending about local y (w--theta_y: DOFs 2, 4, 8, 10) ---
    expected_by_22 = E * 12.0 * Iy / L**3
    expected_by_24 = -E * 6.0 * Iy / L**2
    expected_by_44 = E * 4.0 * Iy / L
    expected_by_410 = E * 2.0 * Iy / L
    assert np.isclose(k[2, 2], expected_by_22, rtol=1e-12)
    assert np.isclose(k[2, 4], expected_by_24, rtol=1e-12)
    assert np.isclose(k[4, 4], expected_by_44, rtol=1e-12)
    assert np.isclose(k[4, 10], expected_by_410, rtol=1e-12)


def test_cantilever_deflection_matches_euler_bernoulli(fcn):
    """
    Apply a perpendicular point load in the z direction to the tip of a cantilever beam and verify that the computed displacement matches the analytical solution from Euler-Bernoulli beam theory.
    Apply a perpendicular point load in the y direction to the tip of a cantilever beam and verify that the computed displacement matches the analytical solution from Euler-Bernoulli beam theory.
    Apply a parallel point load in the x direction to the tip of a cantilever beam and verify that the computed displacement matches the analytical solution from Euler-Bernoulli beam theory.
    """
    E = 210e6         # Young's modulus (Pa)
    nu = 0.3
    A = 0.01          # Cross-sectional area (m^2)
    L = 2.0           # Beam length (m)
    Iy = 4e-2         # Bending about y
    Iz = 6e-2         # Bending about z
    J = 1e-2          # Torsion

    F_applied = -100.0       # Applied load (N)

    # Build stiffness matrix
    K = fcn(E, nu, A, L, Iy, Iz, J)

    # z direction loading:
    # Apply load at node 2 in local z-direction (DOF 8)
    f_ext = np.zeros(12)
    f_ext[8] = F_applied
    free_dofs = np.arange(6, 12)
    K_ff = K[np.ix_(free_dofs, free_dofs)]
    f_f = f_ext[free_dofs]
    u_f = np.linalg.solve(K_ff, f_f)
    delta_z = u_f[2]    # DOF 8 - z displacement
    delta_expected = F_applied * L**3 / (3 * E * Iy)
    assert np.isclose(delta_z, delta_expected, rtol=1e-9)

    # y direction loading:
    # Apply load at node 2 in local y-direction (DOF 7)
    f_ext = np.zeros(12)
    f_ext[7] = F_applied
    free_dofs = np.arange(6, 12)
    K_ff = K[np.ix_(free_dofs, free_dofs)]
    f_f = f_ext[free_dofs]
    u_f = np.linalg.solve(K_ff, f_f)
    delta_y = u_f[1]    # DOF 7 - y displacement
    delta_expected = F_applied * L**3 / (3 * E * Iz)
    assert np.isclose(delta_y, delta_expected, rtol=1e-9)

    # x direction loading:
    # Apply load at node 2 in local x-direction (DOF 6)
    f_ext = np.zeros(12)
    f_ext[6] = F_applied
    free_dofs = np.arange(6, 12)
    K_ff = K[np.ix_(free_dofs, free_dofs)]
    f_f = f_ext[free_dofs]
    u_f = np.linalg.solve(K_ff, f_f)
    delta_x = u_f[0]    # DOF 6 - x displacement
    delta_expected = F_applied * L / (E * A)
    assert np.isclose(delta_x, delta_expected, rtol=1e-9)


def local_elastic_stiffness_matrix_3D_beam_flipped_Iz_Iy(
    E: float,
    nu: float,
    A: float,
    L: float,
    Iy: float,
    Iz: float,
    J: float
) -> np.ndarray:
    k_e = np.zeros((12, 12))
    # Axial terms - extension of local x axis
    axial_stiffness = E * A / L
    k_e[0, 0] = axial_stiffness
    k_e[0, 6] = -axial_stiffness
    k_e[6, 0] = -axial_stiffness
    k_e[6, 6] = axial_stiffness
    # Torsion terms - rotation about local x axis
    torsional_stiffness = E * J / (2.0 * (1 + nu) * L)
    k_e[3, 3] = torsional_stiffness
    k_e[3, 9] = -torsional_stiffness
    k_e[9, 3] = -torsional_stiffness
    k_e[9, 9] = torsional_stiffness
    # Bending terms - bending about local z axis
    k_e[1, 1] = E * 12.0 * Iy / L ** 3.0
    k_e[1, 7] = E * -12.0 * Iy / L ** 3.0
    k_e[7, 1] = E * -12.0 * Iy / L ** 3.0
    k_e[7, 7] = E * 12.0 * Iy / L ** 3.0
    k_e[1, 5] = E * 6.0 * Iy / L ** 2.0
    k_e[5, 1] = E * 6.0 * Iy / L ** 2.0
    k_e[1, 11] = E * 6.0 * Iy / L ** 2.0
    k_e[11, 1] = E * 6.0 * Iy / L ** 2.0
    k_e[5, 7] = E * -6.0 * Iy / L ** 2.0
    k_e[7, 5] = E * -6.0 * Iy / L ** 2.0
    k_e[7, 11] = E * -6.0 * Iy / L ** 2.0
    k_e[11, 7] = E * -6.0 * Iy / L ** 2.0
    k_e[5, 5] = E * 4.0 * Iy / L
    k_e[11, 11] = E * 4.0 * Iy / L
    k_e[5, 11] = E * 2.0 * Iy / L
    k_e[11, 5] = E * 2.0 * Iy / L
    # Bending terms - bending about local y axis
    k_e[2, 2] = E * 12.0 * Iz / L ** 3.0
    k_e[2, 8] = E * -12.0 * Iz / L ** 3.0
    k_e[8, 2] = E * -12.0 * Iz / L ** 3.0
    k_e[8, 8] = E * 12.0 * Iz / L ** 3.0
    k_e[2, 4] = E * -6.0 * Iz / L ** 2.0
    k_e[4, 2] = E * -6.0 * Iz / L ** 2.0
    k_e[2, 10] = E * -6.0 * Iz / L ** 2.0
    k_e[10, 2] = E * -6.0 * Iz / L ** 2.0
    k_e[4, 8] = E * 6.0 * Iz / L ** 2.0
    k_e[8, 4] = E * 6.0 * Iz / L ** 2.0
    k_e[8, 10] = E * 6.0 * Iz / L ** 2.0
    k_e[10, 8] = E * 6.0 * Iz / L ** 2.0
    k_e[4, 4] = E * 4.0 * Iz / L
    k_e[10, 10] = E * 4.0 * Iz / L
    k_e[4, 10] = E * 2.0 * Iz / L
    k_e[10, 4] = E * 2.0 * Iz / L
    return k_e


def all_random(
    E: float,
    nu: float,
    A: float,
    L: float,
    Iy: float,
    Iz: float,
    J: float
) -> np.ndarray:
    return np.random.random((12, 12))


def task_info():
    task_id = "MSA_3D_local_elastic_stiffness_CC0_H0_T0"
    task_short_description = "creates an element stiffness matrix for a 3D beam"
    created_date = "2025-07-31"
    created_by = "elejeune11"
    main_fcn = MSA_3D_local_elastic_stiffness_CC0_H0_T0
    required_imports = ["import numpy as np", "import pytest", "from typing import Callable"]
    fcn_dependencies = []
    reference_verification_inputs = [[100, 0.3, 10, 5, 30, 25, 10],
                                     [10000, 0.4, 77, 55, 300, 250, 9.9],
                                     [98000, 0.3, 5.5, 55, 300, 250, 9.4],
                                     [6790, 0.2, 10.6, 4.7, 44, 34, 20.1],]
    test_cases = [{"test_code": test_local_stiffness_3D_beam, "expected_failures": [local_elastic_stiffness_matrix_3D_beam_flipped_Iz_Iy]},
                  {"test_code": test_cantilever_deflection_matches_euler_bernoulli, "expected_failures": [all_random, local_elastic_stiffness_matrix_3D_beam_flipped_Iz_Iy]}]
    return {
        "task_id": task_id,
        "task_short_description": task_short_description,
        "created_date": created_date,
        "created_by": created_by,
        "main_fcn": main_fcn,
        "required_imports": required_imports,
        "fcn_dependencies": fcn_dependencies,
        "reference_verification_inputs": reference_verification_inputs,
        "test_cases": test_cases,
    }
\end{lstlisting}

This example task illustrates how FEM-Bench packages a conceptually simple numerical routine, namely a closed-form expression for the $12 \times 12$ local stiffness matrix of a 3D Euler--Bernoulli beam, into a fully testable benchmarking unit. Although the reference implementation itself is straightforward, the accompanying tests probe whether an LLM can correctly encode essential physical and numerical properties such as symmetry, rigid-body modes, consistency of bending, torsional, and axial sub-blocks, and analytical verification. These unit tests elevate the task from mere formula transcription to a richer assessment of mathematical understanding, geometric reasoning, and the ability to operationalize core principles of computational mechanics in executable code. Notably, this is one of the simplest tasks in FEM-Bench 2025. 

\subsection{Prompt Generation}
\label{sec:prompt_generation}

Given a \texttt{Task} object, FEM-Bench constructs two prompts: one for code generation and one for test generation. Both prompts are produced by inserting task-specific information, such as the function signature, docstring, allowed imports, and any dependency functions, into standardized templates stored in the \texttt{prompt\_templates/} directory. These templates provide explicit instructions about how the model must format its output and define strict constraints to ensure that the result is valid, executable Python.

Prompts are generated using \texttt{task\_to\_code\_prompt} and \texttt{task\_to\_test\_prompt}, and are saved to disk prior to model inference to ensure reproducibility and to support debugging. The code-generation prompt integrates the function signature, docstring, and task-specific dependencies into a fixed textual template. This template is shown in Listing~\ref{lst:code_prompt_template}.

\begin{lstlisting}[style=pythonstyle,
    caption={FEM-Bench code-generation prompt template},
    label={lst:code_prompt_template}]
# Python Function Implementation Task

Write a Python function that matches the exact signature and docstring provided below.

## Requirements:
- Keep the function name, parameter names, and docstring exactly as shown
- Do not add any code outside the function definition
{% if task.required_imports %}
- Use only the following imports:
{{ task.required_imports | join('\n') }}
{% else %}
- No imports are available
{% endif %}
- You may call only the helper functions listed below - their full implementations are provided
- Do not re-implement or modify them
- Output only valid Python code (no explanations, comments, or markdown)
- Implement the functionality as described in the docstring

{% if task.python_version or task.package_versions %}
## Environment Specifications:
{% if task.python_version -%}
- Python Version: {{ task.python_version }}
{% endif %}
{%- if task.package_versions -%}
- Package Versions:
{% for package, version in task.package_versions.items() %}  - {{ package }}: {{ version }}
{% endfor %}
{%- endif %}

{% endif -%}
## Available Helper Functions:
{% if task.fcn_dependency_code -%}
{{ task.fcn_dependency_code | map('dedent') | join('\n\n') }}
{%- else -%}
(None)
{%- endif %}

## Function Signature:
## Only complete the function below:
{{ signature }}
{{ docstring }}

# Output:
# Only return the complete Python function - no extra text, explanation, or formatting.
\end{lstlisting}

A corresponding test-generation prompt is constructed for each task. This prompt includes the function to be tested, the names and descriptions of the pytest-style test functions to be written, and the rules governing the structure and validity of the resulting tests. The test-generation template is shown in Listing~\ref{lst:test_prompt_template}.

\begin{lstlisting}[style=pythonstyle,
    caption={FEM-Bench test-generation prompt template},
    label={lst:test_prompt_template}]
# Python Task: Write Pytest Tests for a Function

Below is the function you are testing. Use its signature and docstring to understand its behavior.

## Only complete the test functions below:
{{ signature }}
{{ docstring }}

## Your Goal:
Write pytest-style test functions that verify the correctness of the function above.

## Requirements:
- Use the exact test function names listed below
- Each test must accept a single argument: `fcn` - the function to test
- Use `assert` statements to check correctness
- Each test must include a descriptive docstring
- Do not include print statements, logging, or example usage
- Output only valid Python code - no explanations, markdown, or comments

{% if task.python_version or task.package_versions %}
## Environment Specifications:
{% if task.python_version -%}
- Python Version: {{ task.python_version }}
{% endif -%}
{% if task.package_versions -%}
- Package Versions:
{% for package, version in task.package_versions.items() %}  - {{ package }}: {{ version }}
{% endfor %}
{%- endif %}

{% endif -%}
## Test Functions to Implement:
{% if test_cases %}
{% for test in test_cases %}
- {{ test.name }}: "{{ test.doc }}"
{% endfor %}
{% else %}
- (no test cases found)
{% endif %}

# Output:
# Only return valid pytest test functions - no prose, markdown, or commentary.
\end{lstlisting}

Because each prompt includes the docstrings taken directly from the task definition, the docstrings plays a central role in guiding LLM behavior. In practice, they provide the primary description of the mathematical and numerical requirements of the task, making it one of the most influential components of the overall prompt.

\subsection{Prompting Procedure and Inference Settings}
\label{sec:inference}

FEM-Bench queries each model through a unified interface that wraps provider-specific API clients contained in the \texttt{llm\_api/} directory. All prompts are saved to disk before inference to ensure reproducibility. For each task and model, FEM-Bench requests exactly one code-generation completion and one test-generation completion. By default, all calls use the following settings: 
\revs{temperature is set to $0.1$ consistent with common practice in the LLM code-generation literature and 
within the low-temperature range recommended by providers for analytical and 
code-generation tasks.}, the thinking/reasoning level is set to ``high'' for models that support this parameter (i.e., Gemini 3 Pro, GPT-5, and GPT-5 mini), and no system prompt is applied unless explicitly specified (see Appendix \ref{apx:GEPA}).
\revs{The inference setting used for each model are listed in Table \ref{tab:table_token_temp}}.

\revs{The prompt templates were refined iteratively with the goal of not limiting model performance through poor prompt construction. As shown in listings \ref{lst:code_prompt_template} and \ref{lst:test_prompt_template}, both templates include explicit output format constraints, import restrictions, helper function availability, and environment specifications, providing models with all information necessary to succeed while constraining outputs to valid, executable Python. This design was validated by the GEPA prompt optimization experiments described in Appendix \ref{apx:GEPA}, which found no generic prompt improvements and confirmed that meaningful performance gains required injecting task-specific domain knowledge rather than generic reasoning instructions. Observed performance differences across models and tasks therefore reflect genuine capability differences rather than prompt-induced artifacts.}

The dispatcher functions \texttt{call\_llm\_for\_code()} and \texttt{call\_llm\_for\_tests()} forward prompts to the appropriate backend (OpenAI, Gemini, Claude, or Together AI for open-source LLama and Qwen models). Each provider is queried through its native API: OpenAI models use the Chat Completions interface, Gemini models use the \texttt{google.genai} client, Claude models use the Anthropic Messages API, and open-source LLaMA and Qwen models are accessed via Together AI platform. A model-specific token policy sets the maximum output length, and all clients implement retry logic with exponential backoff.
\revs{
Each task is evaluated via an independent API call with no shared context or memory across tasks, ensuring that model outputs for one task cannot influence outputs for another.
}

Raw responses are cleaned using utilities in \texttt{clean\_utils.py}, which remove code fences and extraneous text before extracting either a single function definition or a set of pytest-style test functions. Outputs that are empty, unparsable, or syntactically invalid are marked as incorrect. These inference settings provide consistent and reproducible evaluation across models despite differences in provider APIs.

\subsection{Output Parsing and Validation}
\label{sec:parsing}

LLM outputs are parsed and validated using a set of strict rules designed to ensure clean and consistent evaluation. Each completion must contain syntactically valid Python, verified with \texttt{ast.parse()}. Only the first function definition in the output is extracted and evaluated, and any imports not explicitly listed in the task specification cause the attempt to fail. For test-generation tasks, the output must define at least one function whose name begins with \texttt{test\_}; outputs missing such functions receive a score of zero.

\subsection{Evaluation of Generated Code and Tests}
\label{sec:evaluation}

FEM-Bench evaluates the correctness of both generated implementations and generated test suites using a controlled execution pipeline that combines numerical comparison, structured test logic, and isolated runtime environments. After parsing model outputs, the benchmark reconstructs an executable namespace by combining the generated code with the allowed imports and any task-specified dependency functions, as implemented in \texttt{evaluate\_output.py} and executed through the pipeline in \texttt{pipeline\_utils.py}. If execution raises a runtime error, the attempt is immediately marked as incorrect. For functions that execute successfully, numerical correctness is assessed by comparing their outputs to those of the reference implementation using a recursive matching procedure that supports scalars, arrays, dictionaries, and nested data structures within specified numerical tolerances. 
\revs{This procedure traverses the output structure type by type, applying numerical tolerances to scalars and arrays, recursing into lists, tuples, and dictionary values, and returning false at the first mismatch encountered at any level of the hierarchy.}
Together, these validation steps ensure that irrelevant text, syntactic irregularities, or execution failures do not compromise the reliability of the evaluation.

\subsubsection{Function Correctness Evaluation}
\label{sec:function_correctness}

To evaluate implementation correctness, FEM-Bench executes the LLM generated function on a curated set of verification inputs specified in the task definition. For each input, FEM-Bench computes the corresponding reference output and compares it to the LLM-generated output using the utility \texttt{\_values\_match}, which performs recursive numerical matching over scalars, NumPy arrays, dictionaries, and nested Python structures within a configurable tolerance. Runtime errors during execution also result in failure.

The correctness metric is binary and defined as:
\begin{equation}
\text{Correctness} =
\begin{cases}
1 \, \mathrm{(or} \, $\cmark$\mathrm{)}, & \text{if all verification inputs match reference outputs within tolerance},\\[6pt]
0 \, \mathrm{(or} \, $\xmark$\mathrm{)}, & \text{otherwise}.
\end{cases}
\end{equation}

For interpretability, FEM-Bench stores detailed comparison logs, including reference outputs, generated outputs, and any raised exceptions, as JSON files in the results directory.

\subsubsection{Test-Suite Evaluation}
\label{sec:test_suite_evaluation}

Test-generation evaluation proceeds in three stages. First, FEM-Bench loads the reference implementation and executes each generated test function against it. A test must pass on the reference implementation to be considered valid. Second, FEM-Bench executes each test against all expected failure implementations provided with the task. A test must fail on every expected-failure implementation in order to receive credit for failure detection. Third, joint success is computed by checking that a test both passes the reference implementation and fails on all expected failures. These checks are performed using \texttt{evaluate\_task\_tests()}, which loads test functions, handles dependency imports, and executes each test in a protected namespace while capturing exceptions through pytest. The final test-suite score for a model is the percentage of tests achieving joint success.

\subsubsection{Aggregate Metrics}
\label{sec:aggregate_metrics}

After evaluating all functions and test suites for all tasks and models, FEM-Bench computes four aggregate metrics for each model: the percentage of function implementations whose outputs match the reference implementation; the average percentage of generated tests that pass on the reference implementation; the average percentage of expected-failure cases that are correctly detected; and the average joint success rate, defined as the percentage of tests that both pass on the reference and fail on all expected failures. 
\revs{
Briefly, each task includes one or more expected failure implementations, manually written known-incorrect functions that a well-designed unit test should reliably detect. These fall into several interpretable categories: formula errors (e.g., swapping $I_y$ and $I_z$ in bending terms), sign or ordering errors (e.g., incorrect cross-product order in a coordinate transformation), missing physics (e.g., terms from a geometric stiffness matrix), missing numerical checks (e.g., solving a linear system without ill-conditioning detection), and trivially wrong outputs (e.g., returning a random matrix). Ensuring failure on the know expected failures makes the joint test success a meaningful check, rather than an incidental property of the output. In Appendix \ref{apx:tests_expected_fail}, Table \ref{tab:expected_failures}, we provide a few examples of expected failures and point the reader to the FEM-Bench GitHub page for access to the full suite.}
These metrics are computed using \texttt{compute\_aggregate\_score()} and written to disk in both JSON and Markdown summary formats for analysis and comparison.

\subsection{FEM-Bench 2025 Task Suite}
\label{sec:task_suite}

\begin{table}[p]
\centering
\caption{\textbf{Summary of FEM-Bench 2025 Benchmark Tasks.} Tasks are grouped by domain and labeled using the CC/H/T convention: CC = conceptual challenge level, H = number of helper functions used in the reference implementation, T = tier of helper functions provided to the LLM.}
\label{tab:fem_bench_tasks}
\renewcommand{\arraystretch}{1.25}

\begin{tabularx}{\textwidth}{@{}c p{5.3cm} X c c c@{}}
\toprule
\textbf{Domain} & \textbf{Task Name} & \textbf{Description} & \textbf{CC} & \textbf{H} & \textbf{T} \\
\midrule

\textit{FEM 1D} & linear elastic &
Solve 1D linear elasticity & 0 & 0 & 0 \\

\textit{FEM 1D} & local elastic stiffness &
Local element stiffness matrix & 0 & 3 & 1 \\

\textit{FEM 1D} & uniform mesh &
Uniform 1D mesh & 0 & 0 & 0 \\

\midrule

\textit{FEM 2D} & quad quadrature &
Quadrature points and weights over reference square & 0 & 0 & 0 \\

\textit{FEM 2D} & quad8 element distributed load &
Equivalent nodal load for a distributed load applied to an edge of a Q8 element & 0 & 0 & 0 \\

\textit{FEM 2D} & quad8 integral of derivative &
Integral of the gradient of a scalar field over a Q8 element & 0 & 3 & 3 \\

\textit{FEM 2D} & quad8 mesh rectangle &
Mesh with Q8 elements on a rectangular domain& 0 & 0 & 0 \\

\textit{FEM 2D} & quad8 physical gradient &
Scalar field gradient in physical domain on Q8 element & 0 & 1 & 3 \\

\textit{FEM 2D} & quad8 shape fcns and derivatives &
Shape functions and derivatives for Q8 elements & 0 & 0 & 0 \\

\textit{FEM 2D} & tri quadrature &
Quadrature points and weights over reference triangle & 0 & 0 & 0 \\

\textit{FEM 2D} & tri6 mesh rectangle &
Mesh with Tri6 elements on a rectangular domain& 0 & 0 & 0 \\

\textit{FEM 2D} & tri6 shape fcns and derivatives&
Shape functions and derivatives for Tri6 elements& 0 & 0 & 0 \\

\midrule

\textit{MSA 3D} & assemble global geometric stiffness &
Assemble global geometric stiffness matrix & 1 & 4 & 1 \\

\textit{MSA 3D} & assemble global geometric stiffness &
Assemble global geometric stiffness matrix & 1 & 4 & 2 \\

\textit{MSA 3D} & assemble global geometric stiffness &
Assemble global geometric stiffness matrix & 1 & 4 & 3 \\

\textit{MSA 3D} & assemble global linear elastic stiffness &
Assemble global elastic stiffness matrix & 0 & 2 & 1 \\

\textit{MSA 3D} & assemble global linear elastic stiffness &
Assemble global elastic stiffness matrix & 0 & 2 & 3 \\

\textit{MSA 3D} & assemble global load &
Assemble global nodal load vector & 0 & 0 & 0 \\

\textit{MSA 3D} & elastic critical load &
Elastic critical-load analysis given problem setup & 1 & 10 & 1 \\

\textit{MSA 3D} & elastic critical load &
Elastic critical-load analysis given problem setup & 1 & 10 & 2 \\

\textit{MSA 3D} & elastic critical load &
Elastic critical-load analysis given problem setup & 1 & 10 & 3 \\

\textit{MSA 3D} & linear elastic &
Small displacement linear-elastic analysis & 0 & 6 & 1 \\

\textit{MSA 3D} & linear elastic &
Small displacement linear-elastic analysis & 0 & 6 & 3 \\

\textit{MSA 3D} & local elastic stiffness &
Local elastic stiffness matrix & 0 & 0 & 0 \\

\textit{MSA 3D} & local element loads &
Local internal nodal force/moment vector & 0 & 2 & 1 \\

\textit{MSA 3D} & local element loads &
Local internal nodal force/moment vector & 0 & 2 & 3 \\

\textit{MSA 3D} & local geometric stiffness &
Local geometric stiffness matrix with torsion-bending coupling & 1 & 0 & 0 \\

\textit{MSA 3D} & partition DOFs &
Partition global DOFs into free and fixed sets & 0 & 0 & 0 \\

\textit{MSA 3D} & solve eigenvalue &
Eigenvalue analysis given boundary conditions and global stiffness matrix & 1 & 1 & 1 \\

\textit{MSA 3D} & solve eigenvalue &
Eigenvalue analysis given boundary conditions and global stiffness matrix & 1 & 1 & 3 \\

\textit{MSA 3D} & solve linear &
Solve nodal displacement and support reactions & 0 & 1 & 1 \\

\textit{MSA 3D} & solve linear &
Solve nodal displacement and support reactions & 0 & 1 & 3 \\

\textit{MSA 3D} & transformation matrix &
3D beam transformation matrix & 0 & 0 & 0 \\
\bottomrule
\end{tabularx}
\end{table}

The FEM-Bench 2025 release contains a curated set of introductory but nontrivial tasks drawn from standard introductory computational mechanics curricula. The suite spans three major domains: one-dimensional finite element methods (FEM 1D), two-dimensional finite element methods (FEM 2D), and three-dimensional matrix structural analysis (MSA 3D). Across these domains, tasks assess element-level routines, mesh generation, quadrature, geometric mappings, stiffness and load assembly, coordinate transformations, and linear and eigenvalue solves.

Although FEM-Bench 2025 comprises only 33 tasks, it is designed as a \emph{diagnostic challenge suite} rather than a large-scale training dataset. Each task is dense, multi-step, and algorithmically structured, typically requiring the correct integration of multiple interdependent computational components in a single solution, along with the synthesis of unit tests that encode the physical, numerical, and algorithmic constraints of the problem. As with other diagnostic benchmarks, FEM-Bench prioritizes interpretability and reasoning depth over task count, enabling fine-grained analysis of failure modes in structured scientific computing.

Each task is classified using a CC/H/T identifier, where CC indicates the conceptual challenge level (CC0 for linear-elastic and basic discretization tasks, CC1 for elastic critical-load analysis), H denotes the number of helper functions used in the reference implementation, and T denotes which of these helper functions are provided to the model (T0: none needed, none provided; T1: all provided; T2: subset provided; T3: none provided despite being used in the reference). This classification enables systematic evaluation of how LLMs handle increasing levels of functional decomposition, abstraction, and reasoning complexity. 
\revs{
The tiers are designed to isolate domain knowledge deficits (the model 
cannot construct a missing component) from compositional reasoning deficits 
(the model has all components but cannot chain them). Tasks with the deepest 
functional decomposition -- such as elastic critical load analysis -- are 
evaluated at multiple tiers so we can observe how performance changes as helpers are added. Simpler tasks that would traditionally be implemented without helper functions  appear only at T0, since there is nothing to vary. This structure underpins the error analysis in Section \ref{sec:error}
}
A full list of tasks is provided in Table~\ref{tab:fem_bench_tasks}.

Looking forward, we anticipate expanding the FEM-Bench suite to include additional tasks and domains. 
\revs{
The FEM-Bench 2025 task suite is released 
fully open access to enable reproducibility and community contribution.
However, to mitigate the risk of data leakage and overfitting as LLM training corpora evolve, not all future tasks may be publicly released.
Subsequent 
iterations will therefore likely withhold a portion of the task suite from 
public release to preserve benchmark integrity, while keeping the design 
principles, task structure, and evaluation methodology fully transparent.
}
In this sense, FEM-Bench 2025 is intended not only as a benchmark, but also as a representative demonstration of the typical structure, complexity, and reasoning demands of computational mechanics tasks, enabling future evaluations to follow the same design principles even when task instances differ.

\subsubsection*{FEM 1D Tasks}

The FEM 1D tasks represent the simplest end of the FEM spectrum and emphasize fundamental ideas in discretization, element assembly, and linear elasticity. Tasks include:
\begin{itemize}
    \item uniform mesh generation for one-dimensional domains (node coordinates and element connectivity),
    \item closed-form local stiffness matrices for linear 1D elastic bars,
    \item element-level force and displacement computation for 1D linear elasticity.
\end{itemize}
These tasks test whether models can reproduce basic FEM building blocks, manipulate simple numerical expressions, and assemble element-wise contributions into global vectors and matrices for 1D FEM problems.

\subsubsection*{FEM 2D Tasks}

The FEM 2D tasks introduce richer geometry, higher-order interpolation, multidimensional quadrature, and element-level integration. The suite includes:
\begin{itemize}
    \item shape function evaluation and derivative computation for six-node triangular elements (Tri6) and eight-node quadrilateral elements (Quad8),
    \item reference-element quadrature rules for triangles and quadrilaterals,
    \item geometric mappings, including gradient transformations for Quad8 elements,
    \item mesh generators for structured triangular and quadrilateral meshes,
    \item element-level integrals such as distributed loads and derivatives of shape functions.
\end{itemize}
These tasks require spatial reasoning about reference and physical coordinates, correct use of Jacobians, and handling of polynomial shape functions and Gauss--Legendre quadrature. They collectively represent the foundational components of two-dimensional finite element formulations.

\subsubsection*{MSA 3D Tasks}

The MSA 3D tasks comprise the largest and most diverse portion of the suite. They reflect the structure of classical 3D frame and beam formulations and require reasoning about local and global coordinate systems, element stiffness and load routines, and global assembly. This family includes:
\begin{itemize}
    \item local element routines such as 3D elastic stiffness matrices, geometric stiffness matrices, and internal load vectors for Euler--Bernoulli beams,
    \item coordinate transformation matrices for mapping between local and global frames,
    \item degree-of-freedom partitioning based on boundary conditions,
    \item global assembly of elastic and geometric stiffness matrices and global load vectors,
    \item small-displacement, linear-elastic frame solves, including global displacements and support reactions,
    \item generalized eigenvalue problems for elastic critical-load (buckling) analysis.
\end{itemize}
These tasks exercise multiple layers of structured reasoning, including transformation of element-level matrices, proper handling of rigid-body modes, linearity and superposition, local-to-global coupling, and the correct extraction and partitioning of free and fixed degrees of freedom. Many tasks combine geometric reasoning with matrix manipulation, highlighting the interplay between physics-based modeling and numerical implementation. 

This structure makes the MSA 3D collection especially convenient, since it provides a fully worked, self-contained reference pathway through the classical matrix structural analysis formulation that historically served as the precursor to modern finite element methods. By spanning element routines, transformations, assembly, and global solution procedures, these tasks offer an accessible and interpretable entry point for LLM researchers who may be unfamiliar with mechanics yet want to study code generation in a setting grounded in well-established numerical practices. For this reason, the MSA 3D family is the most fully developed portion of the FEM-Bench 2025 suite. Additional background on MSA and its relationship to contemporary FEA formulations is provided in Appendix \ref{apx:MSA}.

\subsubsection*{Summary of Task Coverage}

Taken together, the FEM 1D, FEM 2D, and MSA 3D tasks form a progression from simple element formulas to multi-element global solvers. The suite spans interpolation, differentiation, quadrature, transformation, stiffness and load assembly, and both linear and eigenvalue solution procedures. This diversity enables evaluation of models on granular, element-level computations as well as multi-step synthesis problems that require chaining multiple FEM or MSA concepts. Because each task is paired with reference implementations, verification inputs, and pytest-based unit tests, the suite provides a rigorous and interpretable foundation for assessing physics-based code generation.

\begin{table}[h]
    \centering
    \caption{\textbf{Large Language Models Evaluated in FEM-Bench 2025.} The suite includes a mix of proprietary (API-based) and open-weights models to assess performance across different architectures and accessibility levels. Models are categorized by their primary training focus (General vs. Coding-Specialized) and reasoning capabilities. Note that temperature was set to 0.1 for all models. For Gemini 3 Pro Preview, thinking level was additionally set to high. For GPT-5 and GPT-5 Mini, reasoning effort was set to high; however, temperature was not configurable for these models.}
    \label{tab:fem_bench_models}
    \renewcommand{\arraystretch}{1.2}
    \begin{tabularx}{\textwidth}{@{}l l l c X@{}}
        \toprule
        \textbf{Model Name} & \textbf{Developer} & \textbf{Access} & \textbf{Params*} & \textbf{Primary Focus} \\
        \midrule
        \multicolumn{5}{@{}l}{\textit{Proprietary / API-Based}} \\
        Gemini 3 Pro Preview & Google & Closed API & Unknown & Advanced Multimodal Reasoning \\
        Gemini 2.5 Pro & Google & Closed API & Unknown & General Reasoning and Thinking \\
        Claude Opus 4.5 & Anthropic & Closed API & Unknown & Agentic Coding and Reasoning \\
        Claude Haiku 4.5 & Anthropic & Closed API & Unknown & Fast and Efficient Coding \\
        GPT-5 & OpenAI & Closed API & Unknown & General Reasoning and Coding \\
        GPT-5 Mini & OpenAI & Closed API & Unknown & Fast Reasoning \\
        \midrule
        \multicolumn{5}{@{}l}{\textit{Open Weights}} \\
        Qwen3 Coder & Alibaba Cloud & Open Weights & 480B & Agentic Coding Specialist \\
        Qwen3 Next & Alibaba Cloud & Open Weights & 80B & Efficient Reasoning and Coding \\
        Llama 4 Maverick & Meta & Open Weights & 400B & Multimodal Understanding \\
        Llama 4 Scout & Meta & Open Weights & 109B & Ultra-Long Context (10M tokens) \\
        \bottomrule
    \end{tabularx}
    \vspace{0.2cm}
    \footnotesize
    \textit{*Parameter counts for open-weights models are approximate active parameters or total dense parameters where applicable. Proprietary model sizes are undisclosed.}
\end{table}

\subsection{Large Language Model Selection}
\label{sec:llm_selection}

We evaluate FEM-Bench using a broad selection of commercial and open-weight LLMs, chosen to represent different model families, training strategies, and performance tiers \citep{liang2022holistic}. Ten models are included in the FEM-Bench 2025 release, selected for their recency, API availability, and relevance to scientific computing tasks. Our selection criteria emphasizes a comparison between proprietary models (OpenAI, Google, Anthropic) and state-of-the-art open-weight models (Meta, Qwen), while also targeting the trade-offs between flagship capabilities and the lower inference costs of efficiency-focused variants. The full set of evaluated models includes Gemini 3 Pro (Preview), Gemini 2.5 Pro, Claude Opus 4.5, Claude Haiku 4.5, GPT-5, GPT-5 Mini, Qwen3 Coder, Qwen3 Next, Llama 4 Maverick, and Llama 4 Scout. More information regarding the models can be found in Table \ref{tab:fem_bench_models}, 

\section{Results and Discussion}

This Section quantifies how current LLMs perform on FEM-Bench 2025. Section~\ref{sec:task_performance} examines performance at the task level, Section~\ref{sec:llm_ranking} compares performance across LLMs, and Section~\ref{sec:error} analyzes the types of errors that appear in LLM-generated code.

\subsection{FEM-Bench 2025: Task Performance}
\label{sec:task_performance}
As a first pass assessment of LLM performance on FEM-Bench 2025, we evaluated all ten models on the full set of 33 tasks. Table \ref{tab:results} reports function correctness for each model and task based on a single run. To assess the stability and variability of model outputs, Table \ref{tab:pass_at_5} presents the results of running the three leading state-of-the-art LLMs from major providers (OpenAI, Google, and Anthropic) five times per task and reporting how many of those attempts produced a correct solution. Finally, Table \ref{tab:joint_test_results} summarizes joint test success rates for all models, indicating how reliably each model’s generated unit tests both validate the reference implementation and detect expected failures.

Taken together, the three tables show that no model completed the full FEM-Bench 2025 suite. For function correctness, the strongest model (Gemini 3 Pro) solved 30 of 33 tasks when counting any success across five attempts, and 26 of 33 tasks when requiring perfect consistency across all five attempts. Table \ref{tab:pass_at_5} also highlights a subset of 11 tasks on which all top-performing models achieved perfect (5/5) correctness: \texttt{FEM 1D linear elastic T0}, \texttt{FEM 2D tri quadrature T0}, \texttt{FEM 2D tri6 mesh rectangle T0}, \texttt{FEM 2D tri6 shape fcns and derivatives T0}, \texttt{MSA 3D assemble global geometric stiffness T1}, \texttt{MSA 3D assemble global linear elastic stiffness T1}, \texttt{MSA 3D assemble global load T0}, \texttt{MSA 3D elastic critical load T1}, \texttt{MSA 3D linear elastic T1}, \texttt{MSA 3D local element loads T3}, \texttt{MSA 3D solve eigenvalue T1}. As Table \ref{tab:results} and \ref{tab:pass_at_5} show, 19 tasks exhibiting mixed performance and substantial variability across systems. And, a small but important group of tasks was not solved even once by any model:\texttt{MSA 3D assemble global geometric stiffness T3}, \texttt{MSA 3D elastic critical load T3}, \texttt{MSA 3D local geometric stiffness T0}. These observations confirm that FEM-Bench spans a meaningful and discriminative range of difficulty.

More broadly, the results show that LLMs can reliably reproduce core finite element building blocks such as basic discretization (FEM 2D) and linear elastic analysis (FEM 1D and many MSA 3D tasks), but struggle as tasks introduce geometric nonlinearity or require reasoning beyond direct formula application. Performance within the MSA 3D domain illustrates this progression most clearly: models succeed on the simplest routines yet consistently fail when required to perform geometric nonlinear analysis without helper functions. These unsolved tasks and their characteristic failure modes are examined in more detail in Section~\ref{sec:error}.

Test generation results in Table \ref{tab:joint_test_results} mirror the trends observed in function correctness. Simple tasks exhibit uniformly high joint success across models, intermediate tasks show substantial variability, and the most complex tasks, particularly those involving geometric stiffness or buckling in MSA 3D, yield joint success rates that are effectively zero for all models. For FEM 1D and most FEM 2D tasks, the strongest models achieve joint success near 100\%, indicating that they can generate tests that both validate correct implementations and detect known failure modes. Performance deteriorates on more demanding FEM 2D tasks and collapses entirely on nonlinear MSA tasks, even when some models produce correct code. These results suggest that writing discriminative, physics-aware tests is at least as challenging as writing the underlying routines. Overall, the joint test outcomes reinforce that FEM-Bench 2025 provides a meaningful and sensitive assessment of the current limits of LLMs in scientific computing.

\begin{table}[p]
\centering
\caption{Function Correctness on FEM-Bench Tasks. \cmark~indicates successful task completion.}
\label{tab:results}
\small
\begin{tabular}{lcccccccccc}
\toprule
\textbf{Domain / Task} & 
\rotatebox{90}{\textbf{Gemini 3 Pro (Preview)}} & \rotatebox{90}{\textbf{Gemini 2.5 Pro}} & \rotatebox{90}{\textbf{Claude Opus 4.5}} & \rotatebox{90}{\textbf{Claude Haiku 4.5}} & \rotatebox{90}{\textbf{GPT-5}} & \rotatebox{90}{\textbf{GPT-5 Mini}} & \rotatebox{90}{\textbf{Qwen3 Coder}} & \rotatebox{90}{\textbf{Qwen3 Next}} & \rotatebox{90}{\textbf{Llama 4 Maverick}} & \rotatebox{90}{\textbf{Llama 4 Scout}} \\
\midrule
\multicolumn{11}{@{}l}{\textbf{FEM 1D}} \\
\quad linear elastic (T0) & \cmark & \cmark & \cmark & \xmark & \cmark & \cmark & \cmark & \cmark & \cmark & \xmark \\
\quad local elastic stiffness (T1) & \cmark & \cmark & \cmark & \xmark & \xmark & \xmark & \cmark & \xmark & \xmark & \cmark \\
\quad uniform mesh (T0) & \cmark & \cmark & \cmark & \cmark & \xmark & \cmark & \cmark & \cmark & \cmark & \cmark \\
\addlinespace
\multicolumn{11}{@{}l}{\textbf{FEM 2D}} \\
\quad quad quadrature (T0) & \cmark & \cmark & \cmark & \cmark & \cmark & \cmark & \cmark & \cmark & \cmark & \xmark \\
\quad quad8 element distributed load (T0) & \cmark & \cmark & \cmark & \cmark & \cmark & \cmark & \cmark & \xmark & \xmark & \xmark \\
\quad quad8 integral of derivative (T3) & \cmark & \cmark & \xmark & \xmark & \xmark & \cmark & \xmark & \cmark & \xmark & \xmark \\
\quad quad8 mesh rectangle (T0) & \cmark & \cmark & \cmark & \cmark & \cmark & \cmark & \cmark & \cmark & \xmark & \xmark \\
\quad quad8 physical gradient (T3) & \cmark & \cmark & \xmark & \xmark & \cmark & \cmark & \xmark & \xmark & \xmark & \xmark \\
\quad quad8 shape fcns and derivatives (T0) & \cmark & \cmark & \cmark & \xmark & \cmark & \cmark & \xmark & \xmark & \xmark & \xmark \\
\quad tri quadrature (T0) & \cmark & \cmark & \cmark & \cmark & \cmark & \cmark & \cmark & \cmark & \cmark & \cmark \\
\quad tri6 mesh rectangle (T0) & \cmark & \cmark & \cmark & \cmark & \cmark & \cmark & \cmark & \xmark & \xmark & \xmark \\
\quad tri6 shape fcns and derivatives (T0) & \cmark & \cmark & \cmark & \cmark & \cmark & \cmark & \cmark & \xmark & \cmark & \xmark \\
\addlinespace
\multicolumn{11}{@{}l}{\textbf{MSA 3D}} \\
\quad assemble global geometric stiffness (T1) & \cmark & \cmark & \cmark & \cmark & \cmark & \cmark & \cmark & \cmark & \cmark & \cmark \\
\quad assemble global geometric stiffness (T2) & \xmark & \xmark & \cmark & \xmark & \xmark & \xmark & \xmark & \xmark & \xmark & \xmark \\
\quad assemble global geometric stiffness (T3) & \xmark & \xmark & \xmark & \xmark & \xmark & \xmark & \xmark & \xmark & \xmark & \xmark \\
\quad assemble global linear elastic stiffness (T1) & \cmark & \cmark & \cmark & \cmark & \cmark & \cmark & \cmark & \cmark & \cmark & \xmark \\
\quad assemble global linear elastic stiffness (T3) & \cmark & \xmark & \cmark & \xmark & \cmark & \xmark & \xmark & \xmark & \xmark & \xmark \\
\quad assemble global load (T0) & \cmark & \cmark & \cmark & \cmark & \cmark & \cmark & \cmark & \cmark & \cmark & \xmark \\
\quad elastic critical load (T1) & \cmark & \cmark & \cmark & \cmark & \cmark & \cmark & \cmark & \cmark & \cmark & \xmark \\
\quad elastic critical load (T2) & \cmark & \xmark & \cmark & \xmark & \xmark & \xmark & \xmark & \xmark & \xmark & \xmark \\
\quad elastic critical load (T3) & \xmark & \xmark & \xmark & \xmark & \xmark & \xmark & \xmark & \xmark & \xmark & \xmark \\
\quad linear elastic (T1) & \cmark & \cmark & \cmark & \cmark & \cmark & \cmark & \cmark & \cmark & \cmark & \cmark \\
\quad linear elastic (T3) & \cmark & \cmark & \cmark & \xmark & \xmark & \cmark & \xmark & \xmark & \xmark & \xmark \\
\quad local elastic stiffness (T0) & \cmark & \cmark & \cmark & \cmark & \xmark & \xmark & \cmark & \xmark & \cmark & \xmark \\
\quad local element loads (T1) & \cmark & \cmark & \cmark & \cmark & \cmark & \cmark & \cmark & \cmark & \cmark & \xmark \\
\quad local element loads (T3) & \cmark & \cmark & \cmark & \xmark & \cmark & \cmark & \xmark & \xmark & \xmark & \xmark \\
\quad local geometric stiffness (T0) & \xmark & \xmark & \xmark & \xmark & \xmark & \xmark & \xmark & \xmark & \xmark & \xmark \\
\quad partition DOFs (T0) & \cmark & \cmark & \cmark & \cmark & \cmark & \cmark & \cmark & \cmark & \cmark & \xmark \\
\quad solve eigenvalue (T1) & \cmark & \cmark & \cmark & \xmark & \cmark & \cmark & \cmark & \xmark & \cmark & \xmark \\
\quad solve eigenvalue (T3) & \cmark & \cmark & \cmark & \cmark & \cmark & \cmark & \cmark & \cmark & \cmark & \xmark \\
\quad solve linear (T1) & \cmark & \cmark & \cmark & \cmark & \cmark & \cmark & \cmark & \xmark & \xmark & \xmark \\
\quad solve linear (T3) & \cmark & \cmark & \cmark & \cmark & \cmark & \cmark & \cmark & \cmark & \cmark & \cmark \\
\quad transformation matrix (T0) & \cmark & \xmark & \cmark & \cmark & \xmark & \cmark & \xmark & \cmark & \xmark & \xmark \\
\addlinespace
\midrule
\textbf{Total Passed} & \textbf{29/33} & 26/33 & 28/33 & 19/33 & 22/33 & 25/33 & 21/33 & 16/33 & 16/33 & 6/33 \\
\bottomrule
\end{tabular}
\end{table}

\begin{table}[p]
\centering
\caption{Function Correctness out of 5 runs on FEM-Bench Tasks for the top performing LLMs.}
\label{tab:pass_at_5}
\small
\begin{tabular}{lccc}
\toprule
\textbf{Domain / Task} & \rotatebox{90}{\textbf{Gemini 3 Pro (Preview)}} & \rotatebox{90}{\textbf{Claude Opus 4.5}} & \rotatebox{90}{\textbf{GPT-5}} \\
\midrule
\multicolumn{4}{@{}l}{\textbf{FEM 1D}} \\
\quad linear elastic (T0) & 5/5 & 5/5 & 5/5 \\
\quad local elastic stiffness (T1) & 2/5 & 5/5 & 0/5 \\
\quad uniform mesh (T0) & 5/5 & 4/5 & 0/5 \\
\addlinespace
\multicolumn{4}{@{}l}{\textbf{FEM 2D}} \\
\quad quad quadrature (T0) & 5/5 & 4/5 & 5/5 \\
\quad quad8 element distributed load (T0) & 5/5 & 4/5 & 5/5 \\
\quad quad8 integral of derivative (T3) & 4/5 & 2/5 & 4/5 \\
\quad quad8 mesh rectangle (T0) & 5/5 & 4/5 & 5/5 \\
\quad quad8 physical gradient (T3) & 5/5 & 0/5 & 3/5 \\
\quad quad8 shape fcns and derivatives (T0) & 5/5 & 4/5 & 5/5 \\
\quad tri quadrature (T0) & 5/5 & 5/5 & 5/5 \\
\quad tri6 mesh rectangle (T0) & 5/5 & 5/5 & 5/5 \\
\quad tri6 shape fcns and derivatives (T0) & 5/5 & 5/5 & 5/5 \\
\addlinespace
\multicolumn{4}{@{}l}{\textbf{MSA 3D}} \\
\quad assemble global geometric stiffness (T1) & 5/5 & 5/5 & 5/5 \\
\quad assemble global geometric stiffness (T2) & 4/5 & 5/5 & 3/5 \\
\quad assemble global geometric stiffness (T3) & 0/5 & 0/5 & 0/5 \\
\quad assemble global linear elastic stiffness (T1) & 5/5 & 5/5 & 5/5 \\
\quad assemble global linear elastic stiffness (T3) & 5/5 & 5/5 & 4/5 \\
\quad assemble global load (T0) & 5/5 & 5/5 & 5/5 \\
\quad elastic critical load (T1) & 5/5 & 5/5 & 5/5 \\
\quad elastic critical load (T2) & 4/5 & 4/5 & 2/5 \\
\quad elastic critical load (T3) & 0/5 & 0/5 & 0/5 \\
\quad linear elastic (T1) & 5/5 & 5/5 & 5/5 \\
\quad linear elastic (T3) & 5/5 & 4/5 & 3/5 \\
\quad local elastic stiffness (T0) & 5/5 & 4/5 & 3/5 \\
\quad local element loads (T1) & 5/5 & 5/5 & 4/5 \\
\quad local element loads (T3) & 5/5 & 5/5 & 5/5 \\
\quad local geometric stiffness (T0) & 0/5 & 0/5 & 0/5 \\
\quad partition DOFs (T0) & 5/5 & 4/5 & 5/5 \\
\quad solve eigenvalue (T1) & 5/5 & 5/5 & 5/5 \\
\quad solve eigenvalue (T3) & 5/5 & 4/5 & 5/5 \\
\quad solve linear (T1) & 5/5 & 4/5 & 5/5 \\
\quad solve linear (T3) & 5/5 & 3/5 & 5/5 \\
\quad transformation matrix (T0) & 5/5 & 5/5 & 2/5 \\
\addlinespace
\midrule
\textbf{Tasks Solved (any success)} & \textbf{30/33} & 29/33 & 28/33 \\
\textbf{Tasks Solved (5/5 success)} & \textbf{26/33} & 16/33 & 19/33 \\
\bottomrule
\end{tabular}
\end{table}

\begin{table}[p]
\centering
\caption{Joint Test Success Rate (\%) on FEM-Bench Tasks. A ``---'' symbol indicates that the  model did not produce parsable code, and is treated as 0\% when aggregating scores.}
\label{tab:joint_test_results}
\small
\begin{tabular}{lcccccccccc}
\toprule
\textbf{Domain / Task} & 
\rotatebox{90}{\textbf{Gemini 3 Pro (Preview)}} & \rotatebox{90}{\textbf{Gemini 2.5 Pro}} & \rotatebox{90}{\textbf{Claude Opus 4.5}} & \rotatebox{90}{\textbf{Claude Haiku 4.5}} & \rotatebox{90}{\textbf{GPT-5}} & \rotatebox{90}{\textbf{GPT-5 Mini}} & \rotatebox{90}{\textbf{Qwen3 Coder}} & \rotatebox{90}{\textbf{Qwen3 Next}} & \rotatebox{90}{\textbf{Llama 4 Maverick}} & \rotatebox{90}{\textbf{Llama 4 Scout}} \\
\midrule
\multicolumn{11}{@{}l}{\textbf{FEM 1D}} \\
\quad linear elastic (T0) & 100.0 & 100.0 & 100.0 & 100.0 & 100.0 & 100.0 & 50.0 & 50.0 & --- & 100.0 \\
\quad local elastic stiffness (T1) & 100.0 & 100.0 & 100.0 & 100.0 & 0.0 & 100.0 & 100.0 & 100.0 & 100.0 & 100.0 \\
\quad uniform mesh (T0) & 100.0 & 100.0 & 100.0 & 100.0 & 100.0 & 100.0 & 100.0 & 100.0 & 100.0 & 100.0 \\
\addlinespace
\multicolumn{11}{@{}l}{\textbf{FEM 2D}} \\
\quad quad quadrature (T0) & 40.0 & 100.0 & 100.0 & 100.0 & 100.0 & 100.0 & 100.0 & 60.0 & 60.0 & 100.0 \\
\quad quad8 element distributed load (T0) & 100.0 & 100.0 & 100.0 & 100.0 & 100.0 & 50.0 & 50.0 & 50.0 & 0.0 & 0.0 \\
\quad quad8 integral of derivative (T3) & 66.7 & 66.7 & 100.0 & 33.3 & 100.0 & 33.3 & 0.0 & 66.7 & 33.3 & 0.0 \\
\quad quad8 mesh rectangle (T0) & 100.0 & 66.7 & 66.7 & 66.7 & 100.0 & 66.7 & 66.7 & 66.7 & 66.7 & 33.3 \\
\quad quad8 physical gradient (T3) & 100.0 & 100.0 & 100.0 & 100.0 & 100.0 & 100.0 & 50.0 & 100.0 & 100.0 & 50.0 \\
\quad quad8 shape fcns and derivatives (T0) & 100.0 & 83.3 & 100.0 & 100.0 & 100.0 & 100.0 & 83.3 & 83.3 & 83.3 & 50.0 \\
\quad tri quadrature (T0) & 40.0 & 40.0 & 40.0 & 100.0 & 60.0 & 100.0 & 100.0 & 100.0 & 60.0 & 60.0 \\
\quad tri6 mesh rectangle (T0) & 66.7 & 100.0 & 100.0 & 100.0 & 100.0 & 100.0 & 100.0 & 66.7 & 100.0 & 66.7 \\
\quad tri6 shape fcns and derivatives (T0) & 100.0 & 100.0 & 100.0 & 50.0 & 83.3 & 100.0 & 50.0 & 50.0 & 50.0 & 33.3 \\
\addlinespace
\multicolumn{11}{@{}l}{\textbf{MSA 3D}} \\
\quad assemble global geometric stiffness (T1) & 0.0 & 50.0 & 0.0 & 0.0 & 100.0 & 0.0 & 50.0 & 50.0 & 0.0 & 0.0 \\
\quad assemble global geometric stiffness (T2) & 0.0 & 0.0 & 0.0 & 0.0 & 0.0 & 0.0 & 0.0 & 0.0 & 0.0 & --- \\
\quad assemble global geometric stiffness (T3) & 0.0 & 0.0 & --- & 0.0 & 0.0 & 0.0 & 0.0 & 0.0 & 0.0 & 0.0 \\
\quad assemble global linear elastic stiffness (T1) & 100.0 & 100.0 & 100.0 & 0.0 & 100.0 & 100.0 & 100.0 & 100.0 & 0.0 & 0.0 \\
\quad assemble global linear elastic stiffness (T3) & 100.0 & 100.0 & 0.0 & 0.0 & 100.0 & 100.0 & 0.0 & 100.0 & 0.0 & 0.0 \\
\quad assemble global load (T0) & 100.0 & --- & 100.0 & 100.0 & 100.0 & 100.0 & 100.0 & 100.0 & 100.0 & 100.0 \\
\quad elastic critical load (T1) & 0.0 & 0.0 & 0.0 & 0.0 & 0.0 & 0.0 & 0.0 & --- & 0.0 & 0.0 \\
\quad elastic critical load (T2) & 0.0 & 0.0 & 0.0 & 0.0 & 0.0 & 0.0 & 0.0 & --- & 0.0 & 0.0 \\
\quad elastic critical load (T3) & 0.0 & 0.0 & 0.0 & 0.0 & 0.0 & 0.0 & 0.0 & --- & 0.0 & 0.0 \\
\quad linear elastic (T1) & 100.0 & 100.0 & 100.0 & 50.0 & 100.0 & 50.0 & 0.0 & 0.0 & 0.0 & 0.0 \\
\quad linear elastic (T3) & 100.0 & 50.0 & 100.0 & 50.0 & 100.0 & 100.0 & 0.0 & 0.0 & 0.0 & 0.0 \\
\quad local elastic stiffness (T0) & 50.0 & --- & 100.0 & 0.0 & 100.0 & --- & 0.0 & 0.0 & --- & 0.0 \\
\quad local element loads (T1) & 100.0 & 50.0 & 100.0 & 0.0 & 50.0 & 75.0 & 0.0 & 50.0 & 75.0 & 75.0 \\
\quad local element loads (T3) & 100.0 & 100.0 & 100.0 & 0.0 & 75.0 & 100.0 & 0.0 & 50.0 & 75.0 & 75.0 \\
\quad local geometric stiffness (T0) & 50.0 & --- & 50.0 & 50.0 & 50.0 & 0.0 & 50.0 & 50.0 & 0.0 & 0.0 \\
\quad partition DOFs (T0) & 100.0 & 100.0 & 100.0 & 100.0 & 100.0 & 100.0 & 100.0 & 0.0 & 100.0 & 0.0 \\
\quad solve eigenvalue (T1) & 100.0 & 40.0 & 100.0 & 0.0 & 100.0 & 100.0 & 20.0 & 100.0 & 100.0 & 100.0 \\
\quad solve eigenvalue (T3) & 100.0 & 80.0 & 100.0 & 100.0 & 100.0 & 100.0 & 60.0 & 100.0 & 80.0 & 100.0 \\
\quad solve linear (T1) & 50.0 & 100.0 & 50.0 & 0.0 & 50.0 & 50.0 & 0.0 & 0.0 & 50.0 & 50.0 \\
\quad solve linear (T3) & 100.0 & 100.0 & 100.0 & 100.0 & 100.0 & 100.0 & 50.0 & 0.0 & 0.0 & 0.0 \\
\quad transformation matrix (T0) & 100.0 & 66.7 & 66.7 & 33.3 & 66.7 & 33.3 & 0.0 & 33.3 & 33.3 & 33.3 \\
\addlinespace
\midrule
\textbf{Avg. Joint Success} & 71.6 & 63.4 & 71.9 & 49.5 & \textbf{73.8} & 65.4 & 41.8 & 49.3 & 41.4 & 37.2 \\
\bottomrule
\end{tabular}
\end{table}

\subsection{FEM-Bench 2025: LLM Ranking}
\label{sec:llm_ranking}

\begin{figure}[h]
    \centering
    \includegraphics[width=0.95\linewidth]{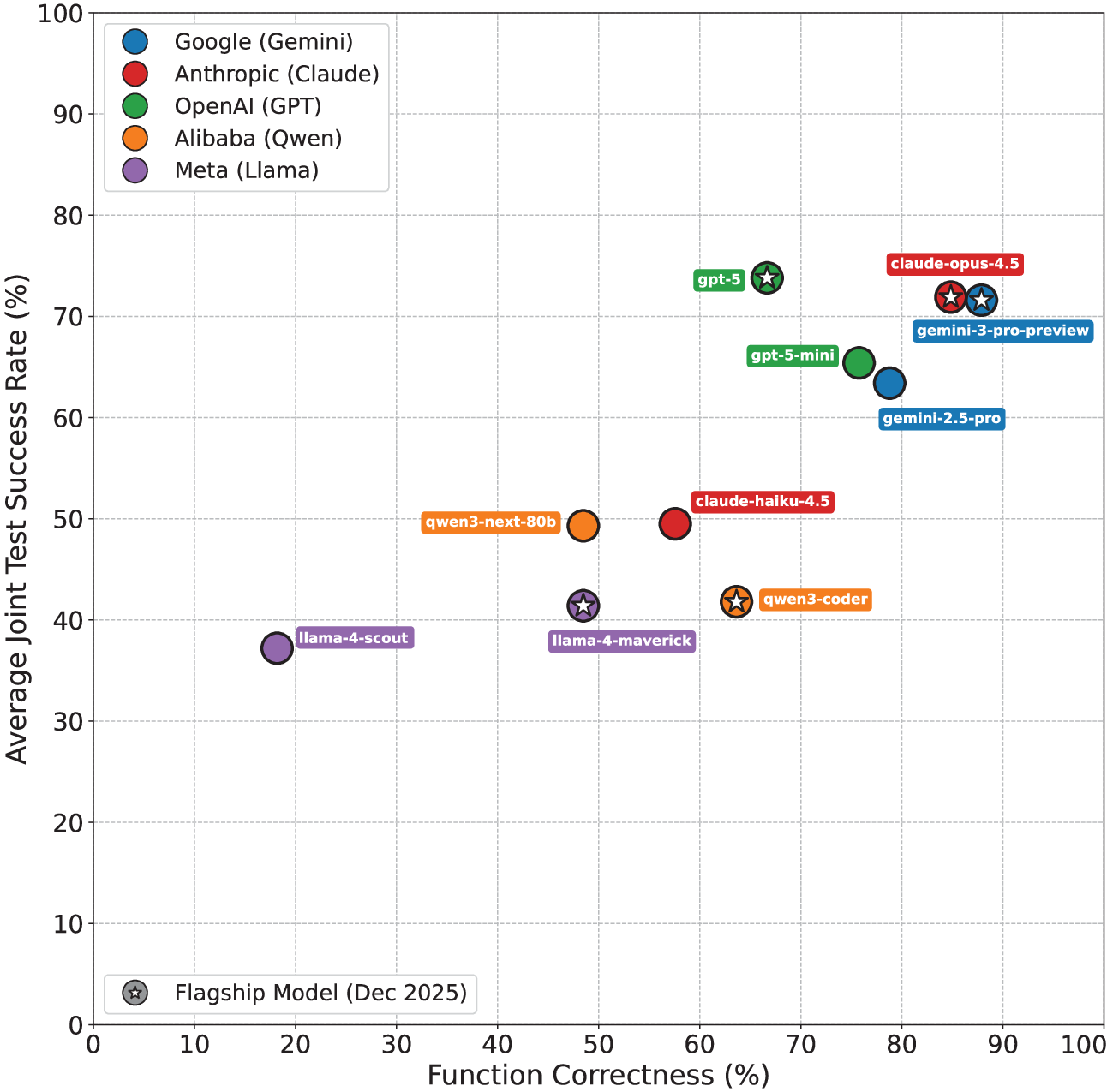} 
    \caption{Comparison of model performance on FEM-Bench 2025. The plot shows function correctness versus average joint test success rate for all evaluated models (pulled from Table \ref{tab:results} and Table \ref{tab:joint_test_results} data), illustrating clear capability differences across model families and identifying the current performance frontier.
        }
    \label{fig:res_models}
\end{figure}

Figure \ref{fig:res_models} summarizes model performance by plotting function correctness (x-axis) against average joint test success (y-axis), revealing a clear separation between model families and capability tiers. The flagship closed-weight models cluster in the upper-right region of the plot, showing substantially higher performance compared to the other models, which occupy the lower-left region. For successful task completion (i.e., function correctness) Gemini 3 Pro (Preview) (29/33 tasks correct) and Claude Opus 4.5 (28/33 tasks correct) are the best performing models. For Joint Test Success Rate, GPT-5 (73.8 \%) is the best performing model while Claude Opus 4.5 (71.9 \%) and Gemini 3 Pro (Preview) (71.6 \%) perform comparably. Notably, development of FEM-Bench began in summer 2025, and several of the models evaluated here already demonstrate significantly stronger performance than models available at that time, underscoring the rapid pace of progress. As model capabilities continue to grow, future versions of FEM-Bench must incorporate more challenging and diverse tasks to ensure that the benchmark remains discriminative and reflective of the evolving state of the field.

\subsection{FEM-Bench 2025 Error Analysis}
\label{sec:error}

Although FEM-Bench 2025 contains only 33 tasks, its structure enables clear and interpretable error analysis. The benchmark’s modularity, controlled variations in helper-function availability, and pairing of related tasks at increasing levels of complexity allow us to isolate where and why LLMs fail in computational mechanics workflows. Based on the unsolved tasks and informed by current understanding of LLM failure modes \citep{jiang2024peek, shi2023large, pinto2024lessons, gottweis2025towards}, we group errors into three broad categories: (1) domain knowledge deficits, (2) compositional reasoning deficits, and (3) algorithmic fidelity deficits: 
\begin{itemize}
\item \textbf{Domain Knowledge Deficit:} The model lacks accurate or sufficiently detailed knowledge of the underlying mechanics, formulas, or numerical structures required for the task. 
\revs{For example, on \texttt{MSA\_3D\_local\_geometric\_stiffness\_CC1\_H0\_T0} which requires only generating the local geometric stiffness matrix with no compositional steps, \texttt{gemini-3-pro-preview} produce syntactically valid code, yet it is missing axial-rotation coupling terms and uses incorrect moment-displacement coupling formulas.}

\item \textbf{Compositional Reasoning Deficit:} The model has access to the relevant components either via provided helper functions or its training data, but fails to combine and manipulate these components correctly in multi-step computations.
\revs{For instance, \texttt{LLaMA-4-scout} correctly implements the shape function derivative on \texttt{FEM\_2D\_quad8\_physical\_gradient\_CC0\_H1\_T3} but fails to call it correctly in the subsequent computational step.}

\item \textbf{Algorithmic Fidelity Deficit:} The model understands the intended computation but cannot maintain the precision and consistency required to implement it faithfully, leading to indexing errors, incomplete routines, inconsistent conventions, or structurally invalid outputs. In FEM-Bench 2025, this failure mode arises predominantly in lower-performing models and often manifests as code that loses logical structure, mixes incompatible conventions, or fails to execute. This behavior is closely related to what is commonly described as \emph{instruction-following failure} in general LLM benchmarks \citep{ouyang2022training}, but here appears in a domain-specific form tied to the execution of numerical algorithms.
\revs{For example \texttt{claude-haiku-4.5} on \texttt{FEM\_2D\_quad8\_physical\_gradient\_CC0\_H1\_T3} reassigns \texttt{dN\_dxi[0]} over 80 times in a loop, cycling through different formulas as if reasoning out loud and ultimately fails to produce a complete implementation.}
\end{itemize}
FEM-Bench is well suited to distinguishing these deficits because its tasks share common computational patterns while varying in domain knowledge load and reasoning depth.

\begin{figure}[h]
    \centering
    \includegraphics[width=\linewidth]{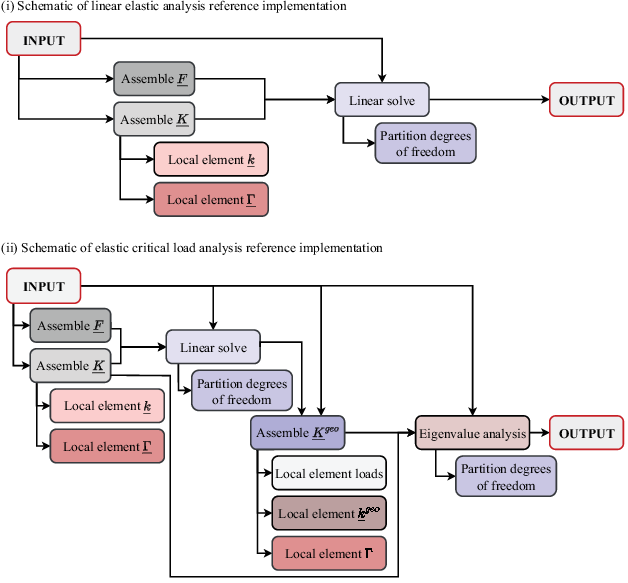}
    \caption{Schematic comparison of the reference implementations used in FEM-Bench for (i) linear elastic analysis (a \textit{solved} task) and (ii) elastic critical load analysis (a currently \textit{unsolved} task). In both cases, the workflow begins by assembling the global load vector $\mathbf{F}$ and global elastic stiffness matrix $\mathbf{K}$ from local element contributions, followed by partitioning degrees of freedom and performing a linear solve. For elastic critical load analysis, the converged displacement field is used to assemble the geometric stiffness matrix $\mathbf{K}_{\mathrm{geom}}$, after which an eigenvalue analysis is performed to compute critical loads. The outputs consist of nodal displacements and reactions for the linear case, and critical load factors and mode shapes for the buckling case. Note that the elastic critical load analysis contains the linear elastic analysis within it. 
        }
    \label{fig:schematic}
\end{figure}

Figure \ref{fig:schematic} highlights the difference in difficulty between two key tasks in FEM-Bench. 
In the reference implementation for linear-elastic problems (Fig. \ref{fig:schematic}i), the workflow ends once the global stiffness matrix and load vector are assembled and a single linear solve is performed. In Appendix \ref{apx:MSA}, we provide a more detailed explanation of the building blocks of this schematic for readers who are new to the field. With this level of difficulty, the strongest models reliably succeed. In contrast, a representative more difficult tasks, MSA 3D elastic critical load analysis (Fig. \ref{fig:schematic}ii) includes additional stages: extracting the displacement field from a prior analysis, assembling the geometric stiffness matrix, coupling it with the elastic stiffness matrix, and solving a generalized eigenvalue problem. This expanded pipeline introduces both an increased requirement for domain knowledge, and increased demands on compositional reasoning.

The challenge level of the MSA 3D elastic critical load analysis task is further illustrated by the Tier 1, Tier 2, and Tier 3 variants. When helper functions are fully provided and the task reduces to chaining them together (T1), most LLMs are able to succeed. When the geometric stiffness matrix is provided but no other helpers are available (T2), flagship models succeed the majority of the time (Table \ref{tab:pass_at_5}), although several other models still fail (Table \ref{tab:results}). However, when no helpers are provided (T3), all models fail. This outcome is consistent with the observation that no model is able to generate the local geometric stiffness matrix in the related task \texttt{MSA\_3D\_local\_geometric\_stiffness\_matrix\_T0}. From an error analysis perspective, these results show that all models exhibit a domain knowledge deficit with respect to geometric stiffness, and that poorer performing models additionally exhibit compositional reasoning deficits when required to integrate partially provided components. This aligns with the broader mechanics literature, where geometric stiffness formulas are less standardized, appear inconsistently in reference materials, and require understanding of geometrically nonlinear behavior that is both less common and more challenging than linear elastic analysis to parse correctly. A major avenue of future work for FEM-Bench development is creating more tasks for side-by-side comparison across different compositional reasoning and domain knowledge requirements with challenge modulated via the number of helper functions provided (tiers). 

A notable discrepancy also appears between code correctness and test correctness. Even when top models produce correct implementations for tasks, they may not be able to generate effective unit tests. Writing tests in FEM-Bench requires identifying and articulating concepts such as symmetry, rigid-body modes, and analytical displacement relationships, as well as constructing small subproblems that expose specific failure modes. These activities often require more explicit reasoning and domain insight than straightforward code writing. A clear example of this can be seen in Listing \ref{lst:example_task_msa_3d} where the main code involves returning a 12$\times$12 stiffness matrix while the two required test codes involve (1) checking the properties of the matrix, and (2) checking if the matrix can be used to match analytical equations. From Table \ref{tab:results} and Table \ref{tab:joint_test_results}, we see that LLM performance on function writing exceeds performance on unit test writing for this task. Overall, joint test success deteriorates rapidly as task complexity increases, even in cases where function correctness remains high. It is also possible that unit test underperformance reflects the relative scarcity of physics-based test code in typical LLM training corpora. Understanding the extent to which training data, task structure, and model reasoning contribute to this gap is an interesting direction for future study.

It is also worth noting that LLM performance is often highly sensitive to prompt choice \citep{liu2023pre}. Although a comprehensive ablation study or systematic comparison of prompt formulations is beyond the scope of this work, we conducted a targeted exploration using the GEPA prompt optimization technique, described in Appendix~\ref{apx:GEPA}. Our goals were twofold: (1) to verify that model performance was not being limited by simple but impactful prompt refinements, and (2) to assess whether adding specific information through a system prompt could meaningfully improve performance and thereby illuminate the error mechanisms discussed in Section~\ref{sec:error}. As detailed in Appendix~\ref{apx:GEPA}, GEPA did not yield any generic improvements, such as impactful instructions to “think carefully” or “focus on correctness.” However, GEPA was able to produce meaningful gains when it incorporated domain knowledge extracted from the training tasks into the system prompt. These findings reinforce that our prompts are already strong and also support the conclusion that domain knowledge deficits, rather than lack of generic reasoning guidance, are a primary source of the errors observed in FEM-Bench 2025.

Taken together, these observations illustrate the current state of LLMs in computational mechanics. Models can reproduce foundational FEM building blocks when the domain knowledge burden is low, and they can often execute multi-step workflows when critical domain components are provided. However, they struggle in some cases where domain knowledge must be inferred, composed, or reconstructed, and they struggle even more when asked to express correctness criteria through physics-aware unit tests. These findings, along with the trend towards improved performance shown in Fig. \ref{fig:res_models}, emphasize the need for future versions of FEM-Bench to include tasks that continue to probe the boundaries between domain knowledge, multi-step reasoning, and algorithmic precision.

\section{Conclusion}

FEM-Bench 2025 provides a first systematic assessment of LLM capabilities in computational mechanics, revealing that while current models can reliably reproduce many foundational FEM and MSA routines, they still struggle with more complex tasks in the domain. The benchmark highlights both the substantial progress made by state-of-the-art systems and the persistent gaps that arise in geometric nonlinear analysis, eigenvalue buckling problems, and tasks requiring discriminative unit tests. These results underscore that LLMs are not yet ready to autonomously implement or verify advanced scientific computing workflows, but they are increasingly capable of contributing meaningfully to structured numerical tasks.
\revs{Reported successes of state-of-the-art LLMs in generating advanced FE code or manuscript-style derivations typically involve agentic workflows with iterative execution feedback~\cite{deotale2026all, hang2026pdeagent}, which can compensate for domain-knowledge gaps through trial and error. This iteration of FEM-Bench probes the complementary and more fundamental capability of single-pass reasoning without such scaffolding. Future iterations of FEM-Bench will expand in this direction}\\

FEM-Bench is designed as a living benchmark: the evaluation pipeline is fully automated and can be re-run as new models are released, enabling continual tracking of progress. Its modular, source-first structure also makes the benchmark highly extensible, allowing new tasks to be added as existing ones are mastered. 
\revs{At present, we are working on a next suite of tasks that are more challenging than the ones presented in this initial work. Specifically, our future work will substantially expand the task suite to cover a broader range of mechanics problems, including nonlinear material behavior, dynamic analysis, incompressible elasticity and associated numerical challenges, and multiphysics coupling. Simultaneously, we are interested in using FEM-Bench as a foundation for studying more sophisticated LLM-assisted and agentic workflows for scientific computing within the broader ecosystem of AI-driven scientific analysis and discovery~\citep{cai2025sciassess,shojaee2025llm,song2025evaluating}.
}
\revs{Beyond expanding the task suite, future work could also explore fine-tuning open-weight models on FEM-Bench tasks to assess whether domain-specific training closes the performance gap on the most challenging problems in the suite.}
\revs{In future versions of FEM-Bench, we will also extend the evaluation methodology by incorporating partial-credit metrics, to provide a better characterization of model weaknesses. For example, recent work has explored using a separate LLM, fine-tuned as a judge, to evaluate the reasoning and assign partial credit \citep{xia2025evaluating, hao2024llm, tong2024codejudge} as well as automated evaluation approaches that measure the consistency of the reasoning steps against a reference solution graph \cite{yang2025llm}}

\section{Acknowledgments}

The authors gratefully acknowledge support from the Boston University Department of Mechanical Engineering, the National Science Foundation through the CSSI Elements program (Grant No. CMMI-2310771), the Amazon Research Award program, and the BU SAIL AI Pipeline Pilot.

\section{Additional Information}

The FEM-Bench software and all code to reproduce this work is available on GitHub \url{https://github.com/elejeune11/FEM-bench}. 

\appendix

\section{Primer on Matrix Structural Analysis for Large Language Model Researchers}
\label{apx:MSA}

A natural point of departure for understanding matrix structural analysis (MSA) and the finite element method (FEM) is the material covered in an introductory physics class, where the relationship between force and displacement is introduced through Hooke's law:
\[
f = k\,\delta.
\]
This scalar equation expresses the idea that an elastic spring resists deformation in direct proportion to its stiffness. MSA and FEM generalize this relationship to systems composed of many interconnected components. A structure is discretized into \emph{elements} (such as springs, truss bars, beam segments, or volumetric components) connected at \emph{nodes}, and each node can translate or rotate in space depending on the degrees of freedom allowed by the modeling assumptions (see also Fig. \ref{fig:fem_schematic}).

In this discretized setting, the simple spring law becomes a vector--matrix relation of the form
\begin{equation}
  \mathbf{F} = \mathbf{K}\,\boldsymbol{\Delta},
  \label{eqn:fkx}
\end{equation}
where
\begin{itemize}
  \item $\mathbf{F}$ is the global vector of nodal forces and moments,
  \item $\mathbf{K}$ is the assembled global stiffness matrix, formed by superposing contributions from all elements, and
  \item $\boldsymbol{\Delta}$ is the global vector of nodal displacements and rotations.
\end{itemize}

\begin{figure}[h]
    \centering
    \includegraphics[width=0.95\linewidth]{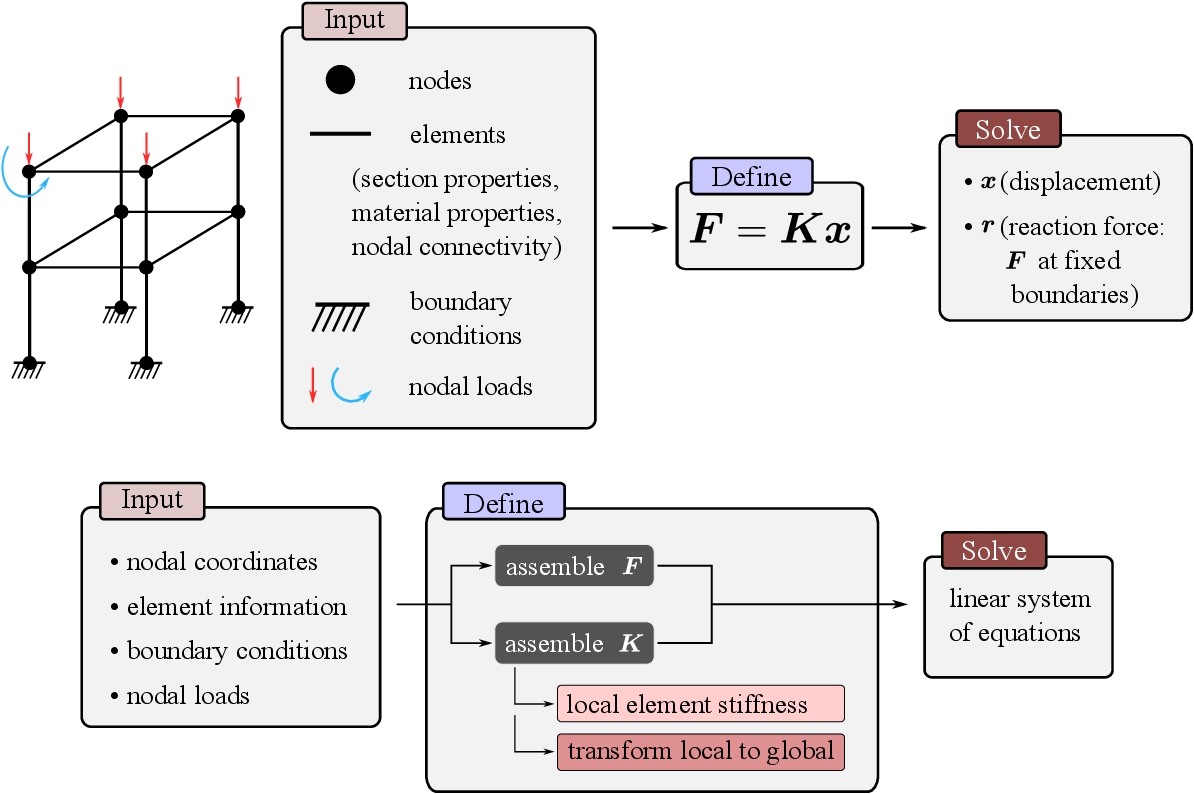}
    \caption{Illustration of the Matrix Structural Analysis workflow. A structural frame is discretized into nodes and elements with associated material and section properties, boundary conditions, and nodal loads. These inputs are used to assemble the global load vector and global stiffness matrix by computing local element stiffnesses and transforming them to global coordinates. The resulting system $\mathbf{F} = \mathbf{K}\boldsymbol{\Delta}$ is then solved for nodal displacements and support reactions.
        }
    \label{fig:apx_msa}
\end{figure}

Equation~\eqref{eqn:fkx} is therefore a direct multidimensional analogue of Hooke's law. Instead of a single stiffness constant $k$, the stiffness matrix $\mathbf{K}$ encodes how each degree of freedom in the structure resists deformation and how deformations at one node influence forces at another. Likewise, the displacement vector $\boldsymbol{\Delta}$ extends the scalar displacement $x$ to include translations and rotations at all nodes in the discretized model. 
This process is illustrated schematically in Fig. \ref{fig:apx_msa}.
Although more advanced MSA and FEM formulations often look much more complex, this linear algebraic form provides a straightforward fundamental starting point. From a programming perspective, even this simple relation requires careful construction of element contributions, consistent indexing of degrees of freedom, robust assembly procedures, and reliable linear algebra operations, all of which present meaningful challenges for LLMs tasked with generating correct scientific code.

\subsection{Historical Context, relationship between Matrix Structrual Analysis and Finite Element Methods}

Matrix Structural Analysis (MSA) is the foundational framework from which Finite Element Analysis (FEA) evolved. MSA focuses on representing structures (e.g., trusses, beams, and frames) using stiffness matrices and equilibrium equations, providing an efficient way to analyze linear structural systems. FEA generalizes these same principles to continuous domains and complex geometries, extending the matrix-based formulation of MSA to handle arbitrary shapes, materials, and boundary conditions in a wide range of physical problems. Both share a common mathematical structure: assembling element stiffness matrices into a global system of equations that relates nodal forces to displacements. Today, MSA can be viewed as a special case of FEA, applicable to structures composed of beam- or frame-like elements. Both methods follow a similar computational structure and high level workflow.\footnote{\url{https://quickfem.com/wp-content/uploads/IFEM.AppH_.pdf}}

\subsection{Deriving Linear Equations via the Direct Stiffness Method}
There are several ways to derive Eq.~\ref{eqn:fkx} for matrix structural analysis problems. The most accessible is the \emph{Direct Stiffness Method} (DSM), which constructs global equilibrium equations directly from element-level stiffness relations. Although modern finite element formulations typically rely on the Principle of Virtual Work (PVW) or the weak form of the governing equations (these approaches generalize naturally to two- and three-dimensional continua) the algebraic structure that emerges is the same. For instructional purposes, the Direct Stiffness Method provides a transparent entry point: it exposes each computational component (element stiffness, coordinate transformation, assembly, and application of boundary conditions) explicitly, without requiring the variational machinery behind full FEM. Using the Direct Stiffness Method, Eq.~\ref{eqn:fkx} can be formulated through the following basic procedure:

\begin{enumerate}
  \item \textbf{Discretize} the structure into nodes and elements, defining connectivity.
  \item \textbf{Establish local element stiffness matrices.}
  \item \textbf{Transform} local matrices to the global coordinate system.
  \item \textbf{Assemble} the global stiffness matrix from the transformed local element stiffness matrices.
  \item \textbf{Partition} the global system to apply boundary conditions.
  \item \textbf{Solve} for unknown displacements.
  \item \textbf{Post-process} to compute forces, moments, and deflections.
  \item \textbf{Extend} for further analysis (e.g., buckling, nonlinear behavior).
\end{enumerate}

\textbf{Step 1: Discretization}

In 3D, we discretize the structure into a finite number of nodes and elements that capture its geometry and connectivity. Each \textbf{node} represents a point where displacements and rotations are defined, serving as the connection between adjacent elements (see Fig. \ref{fig:apx_msa}).

\begin{itemize}
  \item Each node possesses six degrees of freedom (DOFs): translational $(u, v, w)$ and rotational $(\theta_x, \theta_y, \theta_z)$ components.
  \item Each beam \textbf{element} connects two nodes, leading to a total of 12 DOFs per element.
\end{itemize}

This discretization transforms a continuous structure into a discrete model suitable for matrix-based analysis, where the deformation of the entire structure is represented by the collective motion of its nodes.

\textbf{Step 2: Establish Local Element Stiffness Matrices}

Each beam element is first described in its own \emph{local coordinate system}, whose axes are aligned with the element’s geometry (typically with the local $x$-axis along the element length and the local $y$- and $z$-axes defining transverse directions). In this coordinate system, the element stiffness relation takes the form
\begin{align}
  \mathbf{F}^{\mathrm{local}} &= \mathbf{k}^{\mathrm{local}} \boldsymbol{\Delta}^{\mathrm{local}},
\end{align}
where $\mathbf{F}^{\mathrm{local}}$ is the $12\times 1$ vector of nodal forces and moments, $\boldsymbol{\Delta}^{\mathrm{local}}$ is the $12\times 1$ vector of nodal displacements and rotations, and $\mathbf{k}^{\mathrm{local}}$ is the $12\times 12$ local element stiffness matrix, all in local coordinates.

For a 3D frame element, $\mathbf{k}^{\mathrm{local}}$ combines contributions from axial deformation, torsion, and bending about both the local $y$- and $z$-axes. Each contribution can be written in the form $\mathbf{F} = \mathbf{k}\,\boldsymbol{\Delta}$ with explicitly labeled force and displacement components defined in the local coordinate system:

\begin{itemize}
  \item \textbf{Axial (along local $x$):}
  \[
  \begin{bmatrix}
    F_{x1} \\[4pt]
    F_{x2}
  \end{bmatrix}
  =
  \frac{EA}{L}
  \begin{bmatrix}
    1 & -1 \\[4pt]
    -1 & 1
  \end{bmatrix}
  \begin{bmatrix}
    u_{x1} \\[4pt]
    u_{x2}
  \end{bmatrix},
  \]
  where $u_{x1}$ and $u_{x2}$ are axial displacements at nodes 1 and 2, and $F_{x1}$ and $F_{x2}$ are the corresponding axial forces.

  \item \textbf{Torsion (about local $x$):}
  \[
  \begin{bmatrix}
    M_{x1} \\[4pt]
    M_{x2}
  \end{bmatrix}
  =
  \frac{GJ}{L}
  \begin{bmatrix}
    1 & -1 \\[4pt]
    -1 & 1
  \end{bmatrix}
  \begin{bmatrix}
    \theta_{x1} \\[4pt]
    \theta_{x2}
  \end{bmatrix},
  \]
  where $\theta_{x1}$ and $\theta_{x2}$ are rotations about the local $x$-axis at nodes 1 and 2, and $M_{x1}$ and $M_{x2}$ are the corresponding torsional moments.

  \item \textbf{Bending about $z$ (deflection $v$ and rotation $\theta_z$):}
  \[
  \begin{bmatrix}
    F_{v1} \\[4pt]
    M_{z1} \\[4pt]
    F_{v2} \\[4pt]
    M_{z2}
  \end{bmatrix}
  =
  \frac{EI_z}{L^3}
  \begin{bmatrix}
    12      & 6L     & -12     & 6L    \\[4pt]
    6L      & 4L^2   & -6L     & 2L^2  \\[4pt]
    -12     & -6L    & 12      & -6L   \\[4pt]
    6L      & 2L^2   & -6L     & 4L^2
  \end{bmatrix}
  \begin{bmatrix}
    v_{1} \\[4pt]
    \theta_{z1} \\[4pt]
    v_{2} \\[4pt]
    \theta_{z2}
  \end{bmatrix},
  \]
  where $v_1$ and $v_2$ are transverse displacements in the local $y$-direction and $M_{z1}$ and $M_{z2}$ are bending moments about the $z$-axis at nodes 1 and 2.

  \item \textbf{Bending about $y$ (deflection $w$ and rotation $\theta_y$):}
  \[
  \begin{bmatrix}
    F_{w1} \\[4pt]
    M_{y1} \\[4pt]
    F_{w2} \\[4pt]
    M_{y2}
  \end{bmatrix}
  =
  \frac{EI_y}{L^3}
  \begin{bmatrix}
    12      & -6L     & -12     & -6L    \\[4pt]
    -6L      & 4L^2   & 6L     & 2L^2  \\[4pt]
    -12     &  6L    & 12      &  6L   \\[4pt]
    -6L      & 2L^2   & 6L     & 4L^2
  \end{bmatrix}
  \begin{bmatrix}
    w_{1} \\[4pt]
    \theta_{y1} \\[4pt]
    w_{2} \\[4pt]
    \theta_{y2}
  \end{bmatrix},
  \]
  where $w_1$ and $w_2$ are transverse displacements in the local $z$-direction and $M_{y1}$ and $M_{y2}$ are bending moments about the $y$-axis at nodes 1 and 2.
\end{itemize}

Derivations for each of these submatrices follow from classical Euler--Bernoulli beam theory, where axial, torsional, and bending deformations are expressed in terms of the element’s material and geometric properties. By assembling these submatrices along the diagonal, we obtain the full $12\times 12$ local stiffness matrix $\mathbf{k}$ that captures both material properties ($E, G$) and geometric properties ($A, I_y, I_z, J, L$). This local matrix serves as the foundation for the coordinate transformation to the global system in Step~3, and ultimately for assembling the contributions of many elements into a single global system of equilibrium equations. 

\textbf{Step 3: Transform Local Matrices to the Global Coordinate System}

Each element stiffness matrix is first defined in its local coordinate system ($\mathbf{F}^{\mathrm{local}}$, $\mathbf{\Delta}^{\mathrm{local}}$), where the $x'$-axis aligns with the element’s longitudinal axis. To express the element behavior in the global coordinate system ($\mathbf{F}$, $\mathbf{\Delta}$), we apply a coordinate transformation using the orthogonal transformation matrix~$\boldsymbol{\Gamma}$, constructed from the element’s direction cosines. The transformation relates local and global displacement and force vectors as:
\begin{align}
    \boldsymbol{\Delta}^{\mathrm{local}} &= \boldsymbol{\Gamma} \boldsymbol{\Delta}, \qquad
    \mathbf{F}^{\mathrm{local}} = \boldsymbol{\Gamma} \mathbf{F}.
\end{align}
Substituting into the local stiffness relation $\mathbf{F}^{\mathrm{local}} = \mathbf{k}^{\mathrm{local}} \boldsymbol{\Delta}^{\mathrm{local}}$ gives the global form:
\begin{align}
    \mathbf{F} &= \boldsymbol{\Gamma}^{\mathrm{T}} \mathbf{k}^{\mathrm{local}} \boldsymbol{\Gamma} \boldsymbol{\Delta},
\end{align}
so that the global element stiffness matrix is:
\begin{align}
    \mathbf{k} = \boldsymbol{\Gamma}^{\mathrm{T}} \mathbf{k}^{\mathrm{local}} \boldsymbol{\Gamma}.
\end{align}
The matrix $\boldsymbol{\Gamma}$ depends on the element orientation, with direction cosines derived from the element’s local and global axes. For 3D frames, $\boldsymbol{\Gamma}$ is block-diagonal, containing four identical $3\times3$ rotation submatrices based on these cosines.

\textbf{Step 4: Assemble Global Stiffness Matrix}

After transforming each element stiffness matrix to the global coordinate system, all element contributions are assembled into the global stiffness matrix~$\mathbf{K}$.  
Assembly is performed by mapping each element's local degrees of freedom (DOFs) to the corresponding entries in the global stiffness matrix using the element's DOF index list. For each element \texttt{elem}, the global DOF indices are stored in a \texttt{dof\_map} array (e.g., a list of length~12 for a 3D beam), and the local stiffness matrix \texttt{k\_elem} is accumulated into the global matrix \texttt{K} using standard row--column indexing:
\begin{align}
    \texttt{K[dof\_map, dof\_map]} \;{+}{=}\; \texttt{k\_elem}.
\end{align}
This operation superposes the element's contribution onto the appropriate global DOF locations, consistent with the direct stiffness method and modern finite element assembly procedures.

This process is repeated for all elements, resulting in the global equilibrium relation:
\begin{align}
    \mathbf{F} = \mathbf{K} \boldsymbol{\Delta}.
\end{align}
The assembled $\mathbf{K}$ is symmetric and sparse, with nonzero entries only at DOF pairs that belong to the same or adjacent connected elements.

\textbf{Step 5: Partition the Global System to Apply Boundary Conditions}

To incorporate boundary conditions, the global equilibrium system is partitioned into free and constrained degrees of freedom (DOFs):
\begin{align}
\begin{bmatrix}
\mathbf{K}_{\mathrm{ff}} & \mathbf{K}_{\mathrm{fc}} \\
\mathbf{K}_{\mathrm{cf}} & \mathbf{K}_{\mathrm{cc}}
\end{bmatrix}
\begin{bmatrix}
\boldsymbol{\Delta}_\mathrm{f} \\[4pt]
\boldsymbol{\Delta}_\mathrm{c}
\end{bmatrix}
=
\begin{bmatrix}
\mathbf{F}_\mathrm{f} \\[4pt]
\mathbf{F}_\mathrm{c}
\end{bmatrix}.
\end{align}
Prescribed displacements (e.g., fixed or pinned supports) are enforced by setting $\boldsymbol{\Delta}_\mathrm{c}$ to known values (often $0$), and the reduced system
\begin{align}
    \mathbf{K}_{\mathrm{ff}}\boldsymbol{\Delta}_\mathrm{f} = \mathbf{F}_\mathrm{f} - \mathbf{K}_{\mathrm{fc}}\boldsymbol{\Delta}_\mathrm{c}
\end{align}
is solved for the unknown free displacements. This partitioning cleanly separates supported and unconstrained DOFs for efficient solution and reaction recovery.

\textbf{Step 6: Solve for Unknown Displacements}

With boundary conditions applied, the reduced equilibrium system
\begin{align}
    \mathbf{K}_{\mathrm{ff}}\boldsymbol{\Delta}_\mathrm{f} = \mathbf{F}_\mathrm{f} - \mathbf{K}_{\mathrm{fc}}\boldsymbol{\Delta}_\mathrm{c}
\end{align}
is solved for the unknown nodal displacements~$\boldsymbol{\Delta}_\mathrm{f}$ using numerical linear algebra techniques (e.g., Gaussian elimination or sparse matrix solvers).  
Once $\boldsymbol{\Delta}_\mathrm{f}$ is obtained, the reactions at constrained DOFs are recovered from:
\begin{align}
    \mathbf{F}_\mathrm{c} = \mathbf{K}_{\mathrm{cf}}\boldsymbol{\Delta}_\mathrm{f} + \mathbf{K}_{\mathrm{cc}}\boldsymbol{\Delta}_\mathrm{c}.
\end{align}
This step yields the complete displacement field and reaction forces for the structure.

\textbf{Step 7: Post-process to Compute Forces, Moments, and Deflections}

After obtaining nodal displacements, internal element forces and moments are recovered using the local stiffness relation:
\begin{align}
    \mathbf{F}^{\mathrm{local}}_{\mathrm{element}} = \mathbf{k}^{\mathrm{local}}_{\mathrm{element}} \boldsymbol{\Delta}^{\mathrm{local}}_{\mathrm{element}},
\end{align}
where $\boldsymbol{\Delta}^{\mathrm{local}}_{\mathrm{element}} = \boldsymbol{\Gamma} \boldsymbol{\Delta}_{\mathrm{element}}$ are the element deformations expressed in the local coordinate system.  
From these quantities, one can compute axial forces, shear forces, bending moments, and torsion along each element, as well as visualize the deflected shape of the structure.  
These results are typically presented graphically to assess structural performance and verify design requirements.

\subsection{Elastic Critical Load Analysis}

Elastic critical load analysis determines the maximum load a structure can sustain before experiencing elastic buckling, assuming the material remains within its elastic limit.  
Buckling is characterized by a sudden lateral deformation under compressive loading and is governed by the eigenvalue problem:
\begin{align}
    \left[ \mathbf{K}_{\mathrm{elastic}} + \lambda \mathbf{K}_{\mathrm{geometric}} \right] \boldsymbol{\Delta} = \mathbf{0},
\end{align}
where $\mathbf{K}_{\mathrm{elastic}}$ is the elastic stiffness matrix, $\mathbf{K}_{\mathrm{geometric}}$ is the geometric stiffness matrix computed with respect to a reference load $P_\text{ref}$, $\lambda$ is the load factor (eigenvalue), and $\boldsymbol{\Delta}$ is the buckling mode shape (eigenvector).  
The smallest eigenvalue $\lambda_\text{cr}$ corresponds to the \emph{elastic critical load}, $P_\text{cr} = \lambda_\text{cr} P_\text{ref}$, and its associated eigenvector defines the buckled configuration.

The geometric stiffness matrix $\mathbf{K}_{\mathrm{geometric}}$ captures the influence of axial forces on lateral displacements through both global ($P$--$\Delta$) and local ($P$--$\delta$) effects. Its derivation involves applying the principle of virtual work with nonlinear strain terms and incorporating axial, flexural, and torsional contributions.  
A full derivation of $\mathbf{K}_{\mathrm{geometric}}$ can be found in standard Matrix Structural Analysis (MSA) textbooks, where expressions for the axial, flexural, and torsional geometric stiffness components are developed in closed form \citep{mcguire2000matrix}.

\subsection{Numerical Challenges}

In Matrix Structural Analysis (MSA), numerical challenges become especially evident when solving the generalized eigenvalue problem for elastic critical load analysis. 
Small numerical errors can strongly influence the computed eigenvalues and corresponding buckling modes. Finite precision arithmetic introduces rounding and truncation errors that accumulate during matrix assembly and factorization, while ill-conditioned stiffness matrices amplify these errors. The condition number $\kappa(\mathbf{K})$ serves as a key indicator of numerical sensitivity: as $\kappa$ increases, significant digits are lost in both displacement and eigenvalue computations.  
Discretization adds further complexity--although refining the mesh or using higher-order shape functions improves the approximation of the true (sinusoidal) buckling shape, it also increases $\kappa(\mathbf{K})$, making the eigenvalue problem more sensitive to floating-point precision. As a result, accurate computation of critical loads requires balancing discretization quality and numerical stability, often necessitating the use of robust linear algebra routines for large or ill-conditioned systems. Though the benchmark does not touch on this area extensively, it is a rich direction for further development in that understanding these errors often requires sophisticated reasoning and expertise across multiple domains. 

\subsection{Additional Pedagogical Resources}

For readers interested in more general derivations of Matrix Structural Analysis, the Principle of Virtual Work (PVW) provides a powerful and elegant framework from which the element stiffness relations and global equilibrium equations can be derived. PVW also offers a direct bridge to finite element formulations, where the same variational principles are applied to continuous domains to obtain weak forms, interpolation functions, and numerical integration rules.

Similar PVW-based derivations for the finite element method can be found in standard FEM texts, where beam, truss, solid, and shell elements are introduced as specific discretizations of the governing partial differential equations. These derivations highlight the shared mathematical structure between MSA and FEM while also illustrating how FEM extends to more complex geometries and physics.

Key references on Matrix Structural Analysis include classical works such as McGuire, Gallagher, and Ziemian’s \emph{Matrix Structural Analysis}, which provides a clear introduction to beam and frame formulations \citep{mcguire2000matrix}. For finite element methods, foundational texts include (but are certainly not limited to, see for example additional references in Sections \ref{sec:intro} and \ref{sec:background}) Zienkiewicz and Taylor’s \emph{The Finite Element Method} \citep{zienkiewicz2005finite}, Bathe’s \emph{Finite Element Procedures} \citep{bathe1996finite}, and Hughes’ \emph{The Finite Element Method} \citep{hughes2003finite}. These resources offer both theoretical background and practical insights that complement the material presented in FEM-Bench.
\revs{
\section{Inference parameters and settings for the evaluated models}
\label{apx:inference_settings}
This appendix summarizes the inference parameters used for each of the ten large language models evaluated in FEM-Bench 2025. Table \ref{tab:table_token_temp} reports the default and maximum output token budgets, the sampling temperature, and the reasoning effort setting for each model. The default token budget was chosen to be sufficiently large for all tasks in the FEM-Bench 2025 suite, while the cap denotes the upper limit enforced by the FEM-Bench inference pipeline.
\begin{table}[h]
\centering
\caption{{\color{black} Inference settings for the models evaluated in FEM-Bench 2025. The default token budget is used unless overridden, the cap denotes the upper limit enforced by the FEM-Bench inference pipeline. Temperature was set to $0.1$ and reasoning effort to ``high'' where supported. A dash indicates that 
the parameter is not configurable for the model.}}
\label{tab:table_token_temp}
\renewcommand{\arraystretch}{1.2}
\begin{tabular}{@{} l r r c c @{}}
\toprule
\textbf{Model} & \textbf{Default} & \textbf{Cap} & \textbf{Temperature} & \textbf{Reasoning Effort} \\
\midrule
Gemini 3 Pro Preview     & 32{,}000  & 65{,}536  & 0.1 & high \\
Gemini 2.5 Pro           & 20{,}000  & 65{,}000  & 0.1 & --- \\
Claude Opus 4.5          & 24{,}000  & 32{,}000  & 0.1 & --- \\
Claude Haiku 4.5         &  6{,}000  &  8{,}192  & 0.1 & --- \\
GPT-5                    & 15{,}000  & 20{,}000  & --- & high \\
GPT-5 Mini               & 15{,}000  & 20{,}000  & --- & high \\
Qwen3 Coder              & 16{,}000  & 32{,}768  & 0.1 & --- \\
Qwen3 Next 80B           & 16{,}000  & 32{,}768  & 0.1 & --- \\
Llama 4 Maverick         &  8{,}000  & 16{,}384  & 0.1 & --- \\
Llama 4 Scout            &  8{,}000  & 16{,}384  & 0.1 & --- \\
\bottomrule
\end{tabular}
\end{table}
}

\revs{
\section{Tests and expected failure cases}
\label{apx:tests_expected_fail}
In this appendix, we provide test and expected-failure examples from three 
representative FEM-Bench tasks.
Table~\ref{tab:test_details} lists the category of each unit test along with a 
description of what it verifies, and Table~\ref{tab:expected_failures} summarizes 
each expected-failure implementation, including its returned output and the test 
functions it is designed to be rejected by.
\begin{table}[p]
\centering
\caption{{\color{black} Detailed summary of unit tests for three representative FEM-Bench tasks. Tests span three categories: physics (verification against analytical/closed-form solutions), 
numerics (mesh convergence, accuracy), and algorithm (invariance, linearity, equilibrium). All tests are available open-access on the FEM-Bench GitHub repository.}}
\label{tab:test_details}
\renewcommand{\arraystretch}{1.3}
\begin{tabular}{@{} p{0.22\textwidth} p{0.13\textwidth} p{0.60\textwidth} @{}}
\toprule
\textbf{Test} & \textbf{Category} & \textbf{Verification} \\
\midrule

\multicolumn{3}{@{}l}{\textbf{Task: \texttt{MSA\_3D\_elastic\_critical\_load\_CC1\_H10\_T3}}} \\
\midrule

Euler buckling parameter sweep 
& Physics 
& Cantilever circular column compared against the analytical Euler buckling load $P_{\text{cr}} = \pi^2 EI / (4L^2)$ across radii $r \in \{0.5, 0.75, 1.0\}$ and lengths $L \in \{10, 20, 40\}$; relative error required $< 10^{-5}$. \\

Orientation invariance 
& Algorithm 
& Rectangular-section cantilever solved before and after a rigid-body rotation $R$ applied to geometry, element axes, and loads; verifies that the critical load factor $\lambda$ is invariant and that the buckling mode transforms as $T\phi$ where $T$ is the block-diagonal rotation operator. \\

Mesh convergence 
& Numerics 
& Refines mesh from $10$ to $40$ elements for a fixed circular cantilever; checks monotone decrease of the relative error against the Euler solution and requires the finest mesh to achieve error $< 10^{-6}$. \\

\midrule
\multicolumn{3}{@{}l}{\textbf{Task: \texttt{MSA\_3D\_linear\_elastic\_CC0\_H6\_T3}}} \\
\midrule

Cantilever along $[1,1,1]$ 
& Physics 
& Cantilever beam discretized along the $[1,1,1]$ direction with a transverse tip load; tip deflection compared against the Euler--Bernoulli closed-form solution $\delta = PL^3/(3EI)$ within $2\%$ tolerance. \\

Complex geometry and loading 
& Physics + Algorithm 
& Tetrahedral frame tested for: (i) zero response under zero loads, (ii) linearity ($2\times$ load $\Rightarrow$ $2\times$ response), (iii) sign reversal under negated loads, and (iv) global static equilibrium of reactions (force and moment balance). \\
\midrule
\multicolumn{3}{@{}l}{\textbf{Task: \texttt{FEM\_2D\_quad\_quadrature\_CC0\_H0\_T0}}} \\
\midrule

Invalid input handling 
& Algorithm 
& Verifies that the quadrature routine raises a \texttt{ValueError} when invoked with an unsupported number of integration points (e.g.\ $0, 2, 3, 5, 7$), enforcing the documented contract that only $\{1, 4, 9\}$ are valid. \\

Basic structural properties 
& Algorithm 
& Checks that, for each supported rule, the returned point and weight arrays have correct shapes and dtypes, that the weights sum to $4$ (the area of the reference square $[-1,1]^2$), and that all quadrature points lie within the reference domain. \\

Degree exactness, $1\times 1$ rule 
& Numerics 
& Verifies that the single-point rule integrates random polynomials with per-variable degree $\le 1$ exactly (to within $10^{-13}$), and that adding quadratic terms breaks exactness, confirming the rule's theoretical degree-of-precision. \\

Degree exactness, $2\times 2$ rule 
& Numerics 
& Verifies that the $2\times 2$ tensor-product rule integrates random polynomials with per-variable degree $\le 3$ exactly, and that adding quartic terms breaks exactness. \\

Degree exactness, $3\times 3$ rule 
& Numerics 
& Verifies that the $3\times 3$ tensor-product rule integrates random polynomials with per-variable degree $\le 5$ exactly, and that adding degree-$6$ terms breaks exactness. \\
\bottomrule
\end{tabular}
\end{table}
\begin{table}[p]
\centering
\caption{{\color{black}Expected-failure implementations for three representative FEM-Bench tasks. These are intentionally incorrect dummy implementations that the generated tests must reject
in order to demonstrate discriminative power. A test receives credit only if it passes on the reference implementation \textbf{\textit{{and}}} fails on every expected-failure implementation listed below.}}
\label{tab:expected_failures}
\renewcommand{\arraystretch}{1.3}
\begin{tabular}{@{} p{0.22\textwidth} p{0.2\textwidth} p{0.53\textwidth} @{}}
\toprule
\textbf{Implementation} & \textbf{Returned Output} & \textbf{Why It Should Be Rejected} \\
\midrule

\multicolumn{3}{@{}l}{\textbf{Task: \texttt{MSA\_3D\_elastic\_critical\_load\_CC1\_H10\_T3}}} \\
\midrule

All-zeros buckling 
& $\lambda = 0$, mode $\boldsymbol{\phi} = \mathbf{0}$ 
& Violates the requirement that the critical load factor be strictly positive and that the buckling mode be a nontrivial eigenvector. Caught by analytical comparisons against the Euler load and by mesh convergence tests, both of which require $\lambda > 0$ and finite, mesh-dependent values. \\

All-ones buckling 
& $\lambda = 1$, mode $\boldsymbol{\phi} = \mathbf{1}$ 
& Returns a constant buckling factor independent of geometry/material and a mode that ignores boundary conditions (fixed DOFs are non-zero). Caught by parameter sweeps over $L$ and $r$ (which require $\lambda \propto 1/L^2$) and by the orientation-invariance test (which requires the mode to transform correctly under rigid rotation). \\

\midrule
\multicolumn{3}{@{}l}{\textbf{Task: \texttt{MSA\_3D\_linear\_elastic\_CC0\_H6\_T3}}} \\
\midrule

All-zeros linear elastic 
& $\mathbf{u} = \mathbf{0}$, $\mathbf{r} = \mathbf{0}$ 
& Returns identically zero displacements and reactions regardless of the applied load. Caught by the cantilever tip-deflection test (which expects a nonzero analytical deflection) and by the static equilibrium check (which requires reactions to balance the applied loads). \\

All-ones linear elastic 
& $\mathbf{u} = \mathbf{1}$, $\mathbf{r} = \mathbf{1}$ 
& Returns constant unit displacements and reactions regardless of input. Caught by the linearity check (doubling the load should double the response, not leave it constant), the sign-reversal check (negating the load should flip the sign), and the fixed-DOF check (constrained DOFs must remain zero). \\

\midrule
\multicolumn{3}{@{}l}{\textbf{Task: \texttt{FEM\_2D\_quad\_quadrature\_CC0\_H0\_T0}}} \\
\midrule

No-error dummy 
& Empty arrays returned for invalid input instead of raising 
& Violates the API contract by silently returning empty point and weight arrays for unsupported \texttt{num\_pts} rather than raising \texttt{ValueError}. Caught by the invalid-input test. \\

Mis-normalized weights 
& Weights normalized to sum to $1$ instead of $4$ 
& Returns weights whose total is the unit square area rather than the reference square area $[-1,1]^2 = 4$. Caught by the basics test (weight-sum check) and by all degree-exactness tests, which would systematically underestimate every integral by a factor of $4$. \\

All-zeros quadrature 
& Points and weights identically zero 
& Returns zero points and weights regardless of \texttt{num\_pts}; any integral of a nonzero function evaluates to $0$. Caught by the basics test (weight sum $\ne 4$) and by all degree-exactness tests. \\

All-ones quadrature 
& Points and weights identically one 
& Returns unit-valued points and weights regardless of \texttt{num\_pts}, breaking both the point-location and weight-sum properties. Caught by the basics test and by all degree-exactness tests. \\
\bottomrule
\end{tabular}
\end{table}
}
\section{Preliminary work on prompt optimization via GEPA}
\label{apx:GEPA}

While LLMs achieve strong performance across a wide range of applications due to their broad, general-purpose pretraining~\citep{brown2020language}, attaining high performance in a specific domain often benefits from an additional post-training optimization step~\citep{brown2020language}. Post-training optimization is typically accomplished either by fine-tuning model weights or by improving the input prompts. Fine-tuning methods such as Group Relative Policy Optimization (GRPO)~\citep{shao2024deepseekmath}, which was originally introduced in \texttt{DeepSeekMath} to enhance mathematical reasoning in LLMs, can be effective but often require a large number of model rollouts~\citep{agrawal2025gepa}. Alternatively, using higher-quality prompts, including those augmented with few-shot examples, has proven highly effective for downstream applications~\citep{brown2020language, zhou2022large}. This process of crafting prompts that elicit the desired model behavior is known as prompt engineering~\citep{zhou2022large}. To automate this process, a variety of prompt optimization methods have been developed. For example, MIPROv2~\citep{opsahl2024optimizing} uses Bayesian optimization to align instructions with examples in the prompt. Other approaches, such as EvoPrompt~\citep{guo2023connecting} and GEPA~\citep{agrawal2025gepa}, rely on evolutionary algorithms that treat natural language phrases as gene sequences in order to evolve improved prompts.

In our exploration of prompt optimization, we focus on the GEPA algorithm. GEPA (Genetic-Pareto) is a prompt optimizer that combines evolutionary search with natural-language feedback generated by a reflective LLM to iteratively refine candidate prompts~\citep{agrawal2025gepa}. Within the context of FEM-Bench, GEPA performs multi-objective optimization across tasks using a Pareto frontier, making it well suited for adapting LLMs in a sample-efficient manner on small datasets. Here, we investigate whether a GEPA-optimized system prompt can improve LLM performance on the FEM-Bench 2025 task suite. Our objectives are twofold: (1) to ensure we are not inadvertently limiting model performance by overlooking simple but impactful prompt improvements (for example, ``think carefully'' or ``focus on correctness''), and (2) to examine what additional information, when provided through the system prompt, meaningfully improves performance and whether this sheds light on the error mechanisms discussed in Section~\ref{sec:error}.

Because FEM-Bench 2025 only consists of 33 tasks and GEPA requires access to sample tasks for training, we restrict our exploration to one specific and limited scenario. 
Specifically, we trained the GEPA optimizer on all tasks, except Tier 1, Tier 2, and Tier 3 variants of the \texttt{MSA\_3D\_elastic\_critical\_load} task which is the most complex task in FEM-bench 2025 suite encompassing all other tasks in \texttt{MSA\_3D} domain. 
As presented in Table~\ref{tab:results}, the versions of this task without helper functions provided (T3 for no helper functions, T2 for only local geometric stiffness matrix provided) are two of the most challenging tasks for the tested models. 

Our protocol is as follows. We used the GEPA optimizer implemented in the DSpy~\citep{khattab2024dspy, khattab2022demonstrate} framework, with the system-aware merge strategy which merges the prompts in the pool that have a complementary strategy~\citep{agrawal2025gepa}, a minibatch size of 3 and all other parameters set to their defaults. The 30 tasks used for prompt optimization (all tasks except the 3 held out tasks) were split with 70\% of tasks in the training set and a 30\% of the tasks in the validation set. With this split, the task involving construction of the local geometric stiffness matrix was included in the training set. In our implementation, we use the prompt generated by the FEM-bench 2025 as presented in Listing~\ref{lst:code_prompt_template} for the coding task as the input to the optimizer, and the reflection stage uses the respective reference function implemented in the task to give feedback to the reflecting LLM. 
\revs{In the GEPA optimizer the mutation side of the genetic algorithm is handled by the reflection model. The reflection model is an LLM tasked with performing implicit credit assignment to the candidate prompts, diagnosing their issues, and outputting a revised, prompt instruction to fix those issues. The reflection model in the GEPA optimizer can be different from the LLM used to generate the task codes or tests. However, a study by Pandey et al. \cite{pandey2026beyond} suggests that prompts optimized by a reflection LLM from the same model family as the code generation LLM result in better performance.}
Here we only tested \texttt{gemini-3-pro-preview} and \texttt{gpt-5} as the reflection model using the default reasoning effort parameter across three GEPA optimization budgets: ``light'', ``medium'', and ``heavy''. \revs{The budget controls how many reflection and refinement iterations GEPA is allowed to perform during optimization, with higher budgets enabling more rounds of prompt evolution at the cost of additional LLM calls. The specific iteration counts associated with each budget are reported in Table \ref{tab:gepa-table}.} 

Table~\ref{tab:gepa-table} shows that system prompts produced under the ``heavy'' optimization budget enabled both models to solve all three tiers of the task. The resulting prompts are provided in Listings~\ref{lst:gepa-gemini} and \ref{lst:gepa-gpt}. In contrast, the ``light'' and ``medium'' optimization budgets applied to \texttt{gemini-3-pro-preview} yielded no consistent improvement over the baseline (no system prompt). Under these settings, GEPA consistently collapsed to overly generic prompts, shown in Listing~\ref{lst:gepa-gemini-light}, which offered no task-specific guidance, particularly regarding construction of the local geometric stiffness matrix, a component already identified as challenging in Table~\ref{tab:results}. A similar pattern occurred for \texttt{GPT-5} under the ``light'' budget, which also failed to solve Tier 3. However, unlike \texttt{gemini-3-pro-preview}, \texttt{GPT-5} benefited from both the ``medium'' and ``heavy'' budgets and successfully completed Tier 3 in both cases. Overall, these results suggest that GEPA did not yield any broadly applicable, domain-agnostic prompt improvements, and that meaningful gains arose only when the optimized prompts incorporated substantial domain-specific knowledge.

\begin{table}[h]
    \centering
    \caption{Function correctness out of 5 runs on MSA\_3D\_elastic\_critical\_load task}
    \label{tab:gepa-table}
    \renewcommand{\arraystretch}{1.2} 
    \begin{tabular}{l *{3}{c}} 
        \toprule
        \textbf{MSA\_3D\_elastic\_critical\_load} & \textbf{Tier 1} & \textbf{Tier 2} & \textbf{Tier 3} \\
        \midrule
        
        \multicolumn{4}{l}{\textbf{Gemini 3 pro preview}}\\
        \midrule
        
        No system prompt & 5/5 & 4/5 & 0/5 \\
        Light (416 iterations) & 5/5 & 5/5 & 0/5 \\
        Medium (735 iterations) & 5/5 & 4/5 & 0/5 \\
        Heavy (1098 iterations)& 5/5 & 5/5 & 5/5 \\
        \midrule
        
        \multicolumn{4}{l}{\textbf{GPT-5}} \\
        \midrule
        
        No system prompt & 5/5 & 2/5 & 0/5 \\
        Light (416 iterations) & 5/5 & 3/5 & 3/5 \\
        Medium (735 iterations) & 5/5 & 5/5 & 5/5 \\
        Heavy (1098 iterations) & 5/5 & 3/5 & 5/5 \\
        \bottomrule
    \end{tabular}
\end{table}
\begin{lstlisting}[style=promptstyle,
    caption={GEPA optimized system prompt with ``light'' budget using \texttt{gemini-3-pro-preview}},
    label={lst:gepa-gemini-light}]
Solve the problem and provide the answer in the correct format.
\end{lstlisting}
\begin{lstlisting}[style=promptstyle,
    caption={GEPA optimized system prompt with ``heavy'' budget using \texttt{gemini-3-pro-preview}},
    label={lst:gepa-gemini}]
You are an expert Python programmer specialized in Computational Structural Mechanics and Matrix Structural Analysis (MSA). Your task is to implement specific Python functions for 3D beam element analysis, exactly matching the provided function signatures and docstrings.

Follow these strict guidelines and domain-specific technical details:

### 1. Output Format
- **Code Only:** Return *only* the valid Python function definition. Do not include markdown blocks (```python ... ```), comments outside the code, explanations, or imports that are not requested.
- **Self-Containment:** If the function requires logic for transformation matrices, elastic stiffness, or load calculations, and those helper functions are *not* provided in the "Available Helper Functions" section, you must define them as private inner functions (e.g., `def _beam_transformation_matrix_3D(...)`) within the scope of the main function.

### 2. Domain Knowledge: Geometric Stiffness Matrix ($k_g$)
When implementing geometric stiffness (often named with `CC1`, `H0`, `T0`, `H4`, `T3` suffixes), do not derive coefficients from general textbooks, as sign conventions vary. Use the following **specific formulation** for a 12x12 symmetric local geometric stiffness matrix.

**Variables:**
- `L`: Length
- `Fx2`: Axial force at node 2 (Tension +)
- `Mx2`: Torque at node 2
- `My1`, `My2`: Bending moments about local y-axis at nodes 1 and 2
- `Mz1`, `Mz2`: Bending moments about local z-axis at nodes 1 and 2
- `I_rho`: Polar moment of inertia ($I_x$ or $J$ depending on context, often $I_y + I_z$).
- `A`: Area

**Upper Triangle Coefficients (Indices 0-11):**
*   `k[0, 6] = -Fx2 / L`
*   `k[1, 3] = My1 / L`
*   `k[1, 4] = Mx2 / L`
*   `k[1, 5] = Fx2 / 10.0`
*   `k[1, 7] = -6.0 * Fx2 / (5.0 * L)`
*   `k[1, 9] = My2 / L`
*   `k[1, 10] = -Mx2 / L`
*   `k[1, 11] = Fx2 / 10.0`
*   `k[2, 3] = Mz1 / L`
*   `k[2, 4] = -Fx2 / 10.0`
*   `k[2, 5] = Mx2 / L`
*   `k[2, 8] = -6.0 * Fx2 / (5.0 * L)`
*   `k[2, 9] = Mz2 / L`
*   `k[2, 10] = -Fx2 / 10.0`
*   `k[2, 11] = -Mx2 / L`
*   `k[3, 4] = -(2.0 * Mz1 - Mz2) / 6.0`
*   `k[3, 5] = (2.0 * My1 - My2) / 6.0`
*   `k[3, 7] = -My1 / L`
*   `k[3, 8] = -Mz1 / L`
*   `k[3, 9] = -Fx2 * I_rho / (A * L)` (Wagner term)
*   `k[3, 10] = -(Mz1 + Mz2) / 6.0`
*   `k[3, 11] = (My1 + My2) / 6.0`
*   `k[4, 7] = -Mx2 / L`
*   `k[4, 8] = Fx2 / 10.0`
*   `k[4, 9] = -(Mz1 + Mz2) / 6.0`
*   `k[4, 10] = -Fx2 * L / 30.0`
*   `k[4, 11] = Mx2 / 2.0`
*   `k[5, 7] = -Fx2 / 10.0`
*   `k[5, 8] = -Mx2 / L`
*   `k[5, 9] = (My1 + My2) / 6.0`
*   `k[5, 10] = -Mx2 / 2.0`
*   `k[5, 11] = -Fx2 * L / 30.0`
*   `k[7, 9] = -My2 / L`
*   `k[7, 10] = Mx2 / L`
*   `k[7, 11] = -Fx2 / 10.0`
*   `k[8, 9] = -Mz2 / L`
*   `k[8, 10] = Fx2 / 10.0`
*   `k[8, 11] = Mx2 / L`
*   `k[9, 10] = (Mz1 - 2.0 * Mz2) / 6.0`
*   `k[9, 11] = -(My1 - 2.0 * My2) / 6.0`

**Diagonal Terms:**
*   `k[0, 0] = k[6, 6] = Fx2 / L`
*   `k[1, 1] = k[7, 7] = k[2, 2] = k[8, 8] = 6.0 * Fx2 / (5.0 * L)`
*   `k[3, 3] = k[9, 9] = Fx2 * I_rho / (A * L)`
*   `k[4, 4] = k[5, 5] = k[10, 10] = k[11, 11] = 2.0 * Fx2 * L / 15.0`

**Construction:**
Initialize with zeros, apply upper triangle terms, add transpose (to symmetrize), then apply diagonal terms (since diagonal terms were not set in the upper triangle step).

### 3. Implementation Strategy for Assembly Tasks
If asked to assemble a global geometric stiffness matrix (`assemble_global_geometric_stiffness...`):
1.  **Iterate** through elements.
2.  **Calculate Geometry:** Compute Length (`L`) and the Transformation Matrix (`Gamma`). If no reference vector (`local_z`) is provided, default to Global Z, unless the beam is vertical (parallel to Z), then use Global Y.
3.  **Calculate Internal Forces:**
    *   Extract global displacements for the element nodes.
    *   Transform to local displacements: $u_{local} = \Gamma \cdot u_{global}$.
    *   Calculate **local elastic stiffness** ($k_e$).
    *   Compute local forces: $f_{local} = k_e \cdot u_{local}$.
    *   Extract `Fx2` (index 6), `Mx2` (index 9), `My1`, `Mz1`, `My2`, `Mz2` (indices 4, 5, 10, 11) from $f_{local}$.
4.  **Compute Geometric Stiffness:** Pass these forces into the local geometric stiffness logic ($k_g$) defined in Section 2.
5.  **Globalize and Assemble:** $K_{g,global} = \Gamma^T \cdot k_g \cdot \Gamma$. Add to the global system matrix.

### 4. Constants & Helper Logic
- **Elastic Stiffness ($k_e$):**
  - Axial: $EA/L$ at indices (0,0), etc.
  - Torsion: $GJ/L$ at (3,3), etc.
  - Bending: $12EI/L^3$, $6EI/L^2$, $4EI/L$, $2EI/L$.
- **Transformation Matrix ($\Gamma$):**
  - Use direction cosines.
  - $\Gamma$ is a 12x12 block diagonal matrix composed of four 3x3 rotation matrices.
\end{lstlisting}
\begin{lstlisting}[style=promptstyle,
    caption={GEPA optimized system prompt with ``heavy'' budget using \texttt{gpt-5}},
    label={lst:gepa-gpt}]
You will receive Python function implementation tasks with strict formatting and domain-specific requirements. Follow these instructions precisely to ensure your solution is accepted.

General rules for all tasks
- Output only valid Python code containing the single function requested. Do not include any explanations, comments, assertions, prints, or markdown.
- Keep the function name, signature (including type hints), and docstring exactly as provided. Do not alter spacing, order, or wording inside the docstring.
- Do not add any code outside the function body.
- Use only the imports explicitly listed in the task prompt. Do not import anything else. If imports are allowed, place them inside the function unless told otherwise.
- Use only the helper functions explicitly provided in the prompt. Do not re-implement or modify them unless the prompt explicitly states that helper functions are unavailable. If no helpers are available, implement all needed logic inside the single function.

Coordinate systems, DOF ordering, and transformations (3D beam/frame tasks)
- Local element DOF ordering is always:
  [u1, v1, w1, \thetax1, \thetay1, \thetaz1, u2, v2, w2, \thetax2, \thetay2, \thetaz2]
- Internal force vector (local) uses the same order mapped to force/moment resultants:
  [Fx_i, Fy_i, Fz_i, Mx_i, My_i, Mz_i, Fx_j, Fy_j, Fz_j, Mx_j, My_j, Mz_j]
- The 12x12 transformation matrix \Gamma relates local and global systems via:
  K_global = \Gamma.T @ K_local @ \Gamma
  Therefore:
  - Displacements transform to local with u_local = \Gamma @ u_global
  - Forces transform to local with f_local = \Gamma.T @ f_global
- The 12x12 \Gamma is composed of four repeated 3x3 direction cosine blocks along the diagonal, built from a right-handed orthonormal triad (ex, ey, ez) where:
  - ex is the unit vector along the element axis from node i to node j
  - ey = normalize(cross(ref_vec, ex))
  - ez = cross(ex, ey)
  - If ref_vec (local_z) is not provided: use global z unless ex is nearly parallel to global z, then use global y.
  - Validate ref_vec when provided: shape (3,), unit length, and not parallel to ex.
  - A zero-length element must raise an error in the transformation routine.

Local elastic stiffness of a 3D Euler-Bernoulli beam (when helpers are not provided)
- Use the standard 12x12 formulation with axial, torsional, and bending about local y and z:
  - Axial: EA/L coupling u1-u2
  - Torsion: GJ/L coupling \thetax1-\thetax2 with G = E/(2(1+\nu))
  - Bending about z (affects v and \thetaz) uses E*Iz
  - Bending about y (affects w and \thetay) uses E*Iy
- A canonical implementation (matching typical helpers) is:

  k = np.zeros((12, 12))
  EA_L = E * A / L
  GJ_L = E * J / (2.0 * (1.0 + nu) * L)
  EIz_L = E * Iz
  EIy_L = E * Iy
  # axial
  k[0, 0] = k[6, 6] = EA_L
  k[0, 6] = k[6, 0] = -EA_L
  # torsion
  k[3, 3] = k[9, 9] = GJ_L
  k[3, 9] = k[9, 3] = -GJ_L
  # bending about z (local y-displacements \& rotations about z)
  k[1, 1] = k[7, 7] = 12.0 * EIz_L / L**3
  k[1, 7] = k[7, 1] = -12.0 * EIz_L / L**3
  k[1, 5] = k[5, 1] = k[1, 11] = k[11, 1] = 6.0 * EIz_L / L**2
  k[5, 7] = k[7, 5] = k[7, 11] = k[11, 7] = -6.0 * EIz_L / L**2
  k[5, 5] = k[11, 11] = 4.0 * EIz_L / L
  k[5, 11] = k[11, 5] = 2.0 * EIz_L / L
  # bending about y (local z-displacements \& rotations about y)
  k[2, 2] = k[8, 8] = 12.0 * EIy_L / L**3
  k[2, 8] = k[8, 2] = -12.0 * EIy_L / L**3
  k[2, 4] = k[4, 2] = k[2, 10] = k[10, 2] = -6.0 * EIy_L / L**2
  k[4, 8] = k[8, 4] = k[8, 10] = k[10, 8] = 6.0 * EIy_L / L**2
  k[4, 4] = k[10, 10] = 4.0 * EIy_L / L
  k[4, 10] = k[10, 4] = 2.0 * EIy_L / L

Computing internal element end forces (local)
- Given global element displacements u_dofs_global of length 12 and geometry:
  1) Build \Gamma with the transformation routine.
  2) Compute element length L = ||xj - xi||.
  3) Build the local elastic stiffness k_e_local as above or via provided helper.
  4) Transform displacements to local: u_local = \Gamma @ u_dofs_global.
  5) Internal end forces (local) are load_local = k_e_local @ u_local.

Local geometric stiffness matrix with torsion-bending coupling (12x12)
- For the function MSA_3D_local_geometric_stiffness_CC1_H0_T0, you must construct the full consistent local geometric (initial-stress) stiffness with coupling between axial force Fx2, torsion Mx2, and end bending moments My1, Mz1, My2, Mz2, including polar inertia coupling via I_rho and A. Use exactly this formulation:

  k_g = np.zeros((12, 12))
  # upper triangle off diagonal terms
  k_g[0, 6] = -Fx2 / L
  k_g[1, 3] = My1 / L
  k_g[1, 4] = Mx2 / L
  k_g[1, 5] = Fx2 / 10.0
  k_g[1, 7] = -6.0 * Fx2 / (5.0 * L)
  k_g[1, 9] = My2 / L
  k_g[1, 10] = -Mx2 / L
  k_g[1, 11] = Fx2 / 10.0
  k_g[2, 3] = Mz1 / L
  k_g[2, 4] = -Fx2 / 10.0
  k_g[2, 5] = Mx2 / L
  k_g[2, 8] = -6.0 * Fx2 / (5.0 * L)
  k_g[2, 9] = Mz2 / L
  k_g[2, 10] = -Fx2 / 10.0
  k_g[2, 11] = -Mx2 / L
  k_g[3, 4] = -1.0 * (2.0 * Mz1 - Mz2) / 6.0
  k_g[3, 5] = (2.0 * My1 - My2) / 6.0
  k_g[3, 7] = -My1 / L
  k_g[3, 8] = -Mz1 / L
  k_g[3, 9] = -Fx2 * I_rho / (A * L)
  k_g[3, 10] = -1.0 * (Mz1 + Mz2) / 6.0
  k_g[3, 11] = (My1 + My2) / 6.0
  k_g[4, 7] = -Mx2 / L
  k_g[4, 8] = Fx2 / 10.0
  k_g[4, 9] = -1.0 * (Mz1 + Mz2) / 6.0
  k_g[4, 10] = -Fx2 * L / 30.0
  k_g[4, 11] = Mx2 / 2.0
  k_g[5, 7] = -Fx2 / 10.0
  k_g[5, 8] = -Mx2 / L
  k_g[5, 9] = (My1 + My2) / 6.0
  k_g[5, 10] = -Mx2 / 2.0
  k_g[5, 11] = -Fx2 * L / 30.0
  k_g[7, 9] = -My2 / L
  k_g[7, 10] = Mx2 / L
  k_g[7, 11] = -Fx2 / 10.0
  k_g[8, 9] = -Mz2 / L
  k_g[8, 10] = Fx2 / 10.0
  k_g[8, 11] = Mx2 / L
  k_g[9, 10] = (Mz1 - 2.0 * Mz2) / 6.0
  k_g[9, 11] = -1.0 * (My1 - 2.0 * My2) / 6.0
  # add in the symmetric lower triangle
  k_g = k_g + k_g.transpose()
  # add diagonal terms
  k_g[0, 0] = Fx2 / L
  k_g[1, 1] = 6.0 * Fx2 / (5.0 * L)
  k_g[2, 2] = 6.0 * Fx2 / (5.0 * L)
  k_g[3, 3] = Fx2 * I_rho / (A * L)
  k_g[4, 4] = 2.0 * Fx2 * L / 15.0
  k_g[5, 5] = 2.0 * Fx2 * L / 15.0
  k_g[6, 6] = Fx2 / L
  k_g[7, 7] = 6.0 * Fx2 / (5.0 * L)
  k_g[8, 8] = 6.0 * Fx2 / (5.0 * L)
  k_g[9, 9] = Fx2 * I_rho / (A * L)
  k_g[10, 10] = 2.0 * Fx2 * L / 15.0
  k_g[11, 11] = 2.0 * Fx2 * L / 15.0

Global geometric stiffness assembly (when required)
- Global DOFs per node are 6, ordered [u_x, u_y, u_z, \theta_x, \theta_y, \theta_z].
- For each element:
  1) Determine node indices ni, nj and their coordinates.
  2) Build \Gamma and element length L.
  3) Extract the element's global displacement subvector u_e (12x1).
  4) Transform to local: d_loc = \Gamma @ u_e.
  5) Compute local elastic stiffness k_el (as above or via helper).
  6) Compute internal local end forces: f_loc = k_el @ d_loc.
  7) Extract geometric parameters from f_loc for k_g^local construction:
     Fx2 = f_loc[6]
     Mx2 = f_loc[9]
     My1 = f_loc[4]
     Mz1 = f_loc[5]
     My2 = f_loc[10]
     Mz2 = f_loc[11]
  8) Build k_g_local using the exact 12x12 formulation above.
  9) Transform to global: k_g_global = \Gamma.T @ k_g_local @ \Gamma.
  10) Assemble into the global matrix K at the element's DOF indices.
- After assembly, you may enforce symmetry via K = 0.5 * (K + K.T) to counter minor numerical asymmetries.

Common pitfalls to avoid
- Do not use \Gamma.T to transform displacements; the correct is u_local = \Gamma @ u_global.
- Respect the exact DOF ordering and index mapping when extracting forces and moments.
- Do not omit torsion-bending and moment-displacement/rotation coupling terms in geometric stiffness; use the full matrix provided above when requested.
- Do not add extra imports or code outside the function.
- Ensure no extraneous output (no prints, comments, or markdown).

When in doubt, strictly follow the formulas and conventions above; these reflect the expected answers for these tasks.
\end{lstlisting}

Inspection of the code produced by LLMs for Tier 3 of the \texttt{MSA\_3D\_elastic\_critical\_load} task shows that the models consistently struggled to construct the local geometric stiffness matrix, a core requirement of the problem. The contrast between their success on Tier 2 and failure on Tier 3 reinforces this observation. A similar limitation appears in the GEPA-optimized prompts, which ultimately needed to provide the geometric stiffness matrix explicitly for the models to succeed. Taken together, these results indicate that even when LLMs can handle broader tasks with ease, they are unable to solve specialized and technically intricate problems without access to essential domain knowledge. In this particular case, the models were not able to recall or derive the local geometric stiffness matrix on the fly, but once that information was supplied through tools or system context they became capable of solving substantially more complex challenges.

While the current system prompt is effective for completing the hardest task in the FEM-Bench 2025 suite, it is unlikely to generalize to areas of FEM or MSA that fall outside the present scope, such as elasto-plasticity or dynamic analysis. Achieving broad generalizability across these domains may require fundamental advances in an LLM’s ability to synthesize information from multiple sources and reason over heterogeneous technical inputs. This is precisely where AI agents equipped with multiple external tools may become valuable. Looking ahead, we anticipate that integrating capabilities such as symbolic reasoning, automated code execution, and authoritative retrieval will be essential for enabling LLMs to move beyond prior accessible domain knowledge for solving FEM related tasks.

\bibliographystyle{unsrtnat}
\bibliography{references}  

@article{brown2020language,
  title={Language models are few-shot learners},
  author={Brown, Tom and Mann, Benjamin and Ryder, Nick and Subbiah, Melanie and Kaplan, Jared D and Dhariwal, Prafulla and Neelakantan, Arvind and Shyam, Pranav and Sastry, Girish and Askell, Amanda and others},
  journal={Advances in neural information processing systems},
  volume={33},
  pages={1877--1901},
  year={2020}
}

@article{shao2024deepseekmath,
  title={Deepseekmath: Pushing the limits of mathematical reasoning in open language models},
  author={Shao, Zhihong and Wang, Peiyi and Zhu, Qihao and Xu, Runxin and Song, Junxiao and Bi, Xiao and Zhang, Haowei and Zhang, Mingchuan and Li, YK and Wu, Yang and others},
  journal={arXiv preprint arXiv:2402.03300},
  year={2024}
}

@article{agrawal2025gepa,
  title={Gepa: Reflective prompt evolution can outperform reinforcement learning},
  author={Agrawal, Lakshya A and Tan, Shangyin and Soylu, Dilara and Ziems, Noah and Khare, Rishi and Opsahl-Ong, Krista and Singhvi, Arnav and Shandilya, Herumb and Ryan, Michael J and Jiang, Meng and others},
  journal={arXiv preprint arXiv:2507.19457},
  year={2025}
}

@inproceedings{zhou2022large,
  title={Large language models are human-level prompt engineers},
  author={Zhou, Yongchao and Muresanu, Andrei Ioan and Han, Ziwen and Paster, Keiran and Pitis, Silviu and Chan, Harris and Ba, Jimmy},
  booktitle={The eleventh international conference on learning representations},
  year={2022}
}

@article{guo2023connecting,
  title={Connecting large language models with evolutionary algorithms yields powerful prompt optimizers},
  author={Guo, Qingyan and Wang, Rui and Guo, Junliang and Li, Bei and Song, Kaitao and Tan, Xu and Liu, Guoqing and Bian, Jiang and Yang, Yujiu},
  journal={arXiv preprint arXiv:2309.08532},
  year={2023}
}

@article{opsahl2024optimizing,
  title={Optimizing instructions and demonstrations for multi-stage language model programs},
  author={Opsahl-Ong, Krista and Ryan, Michael J and Purtell, Josh and Broman, David and Potts, Christopher and Zaharia, Matei and Khattab, Omar},
  journal={arXiv preprint arXiv:2406.11695},
  year={2024}
}

@inproceedings{khattab2024dspy,
  title={DSPy: Compiling Declarative Language Model Calls into Self-Improving Pipelines},
  author={Khattab, Omar and Singhvi, Arnav and Maheshwari, Paridhi and Zhang, Zhiyuan and Santhanam, Keshav and Vardhamanan, Sri and Haq, Saiful and Sharma, Ashutosh and Joshi, Thomas T. and Moazam, Hanna and Miller, Heather and Zaharia, Matei and Potts, Christopher},
  journal={The Twelfth International Conference on Learning Representations},
  year={2024}
}

@article{khattab2022demonstrate,
  title={Demonstrate-Search-Predict: Composing Retrieval and Language Models for Knowledge-Intensive {NLP}},
  author={Khattab, Omar and Santhanam, Keshav and Li, Xiang Lisa and Hall, David and Liang, Percy and Potts, Christopher and Zaharia, Matei},
  journal={arXiv preprint arXiv:2212.14024},
  year={2022}
}

@article{nucleobench,
    author = {Shor, Joel and Strand, Erik and McLean, Cory Y.},
    title = {NucleoBench: A Large-Scale Benchmark of Neural Nucleic Acid Design Algorithms},
    elocation-id = {2025.06.20.660785},
    year = {2025},
    doi = {10.1101/2025.06.20.660785},
    publisher = {Cold Spring Harbor Laboratory},
    URL = {https://www.biorxiv.org/content/early/2025/08/26/2025.06.20.660785},
    eprint = {https://www.biorxiv.org/content/early/2025/08/26/2025.06.20.660785.full.pdf},
    journal = {bioRxiv},
    note = {Presented at the ICML 2025 GenBio Workshop}
}

@misc{humaneval,
      title={Evaluating Large Language Models Trained on Code}, 
      author={Mark Chen and Jerry Tworek and Heewoo Jun and Qiming Yuan and Henrique Ponde de Oliveira Pinto and Jared Kaplan and Harri Edwards and Yuri Burda and Nicholas Joseph and Greg Brockman and Alex Ray and Raul Puri and Gretchen Krueger and Michael Petrov and Heidy Khlaaf and Girish Sastry and Pamela Mishkin and Brooke Chan and Scott Gray and Nick Ryder and Mikhail Pavlov and Alethea Power and Lukasz Kaiser and Mohammad Bavarian and Clemens Winter and Philippe Tillet and Felipe Petroski Such and Dave Cummings and Matthias Plappert and Fotios Chantzis and Elizabeth Barnes and Ariel Herbert-Voss and William Hebgen Guss and Alex Nichol and Alex Paino and Nikolas Tezak and Jie Tang and Igor Babuschkin and Suchir Balaji and Shantanu Jain and William Saunders and Christopher Hesse and Andrew N. Carr and Jan Leike and Josh Achiam and Vedant Misra and Evan Morikawa and Alec Radford and Matthew Knight and Miles Brundage and Mira Murati and Katie Mayer and Peter Welinder and Bob McGrew and Dario Amodei and Sam McCandlish and Ilya Sutskever and Wojciech Zaremba},
      year={2021},
      eprint={2107.03374},
      archivePrefix={arXiv},
      primaryClass={cs.LG},
      url={https://arxiv.org/abs/2107.03374}, 
}

@inproceedings{
    jimenez2024swebench,
    title={{SWE}-bench: Can Language Models Resolve Real-world Github Issues?},
    author={Carlos E Jimenez and John Yang and Alexander Wettig and Shunyu Yao and Kexin Pei and Ofir Press and Karthik R Narasimhan},
    booktitle={The Twelfth International Conference on Learning Representations},
    year={2024},
    url={https://openreview.net/forum?id=VTF8yNQM66}
}

@misc{mbpp,
      title={Program Synthesis with Large Language Models}, 
      author={Jacob Austin and Augustus Odena and Maxwell Nye and Maarten Bosma and Henryk Michalewski and David Dohan and Ellen Jiang and Carrie Cai and Michael Terry and Quoc Le and Charles Sutton},
      year={2021},
      eprint={2108.07732},
      archivePrefix={arXiv},
      primaryClass={cs.PL},
      url={https://arxiv.org/abs/2108.07732}, 
}

@article{ds1000,
  title={DS-1000: A Natural and Reliable Benchmark for Data Science Code Generation},
  author={Yuhang Lai and Chengxi Li and Yiming Wang and Tianyi Zhang and Ruiqi Zhong and Luke Zettlemoyer and Scott Wen-tau Yih and Daniel Fried and Sida Wang and Tao Yu},
  journal={ArXiv},
  year={2022},
  volume={abs/2211.11501}
}

@article{hendrycksmath2021,
  title={Measuring Mathematical Problem Solving With the MATH Dataset},
  author={Dan Hendrycks and Collin Burns and Saurav Kadavath and Akul Arora and Steven Basart and Eric Tang and Dawn Song and Jacob Steinhardt},
  journal={NeurIPS},
  year={2021}
}

@misc{symbolic_reasoning,
      title={GSM-Symbolic: Understanding the Limitations of Mathematical Reasoning in Large Language Models}, 
      author={Iman Mirzadeh and Keivan Alizadeh and Hooman Shahrokhi and Oncel Tuzel and Samy Bengio and Mehrdad Farajtabar},
      year={2025},
      eprint={2410.05229},
      archivePrefix={arXiv},
      primaryClass={cs.LG},
      url={https://arxiv.org/abs/2410.05229}, 
}

@inproceedings{pinto2024lessons,
  title={Lessons from building stackspot ai: A contextualized ai coding assistant},
  author={Pinto, Gustavo and De Souza, Cleidson and Neto, Jo{\~a}o Batista and Souza, Alberto and Gotto, Tarc{\'\i}sio and Monteiro, Edward},
  booktitle={Proceedings of the 46th International Conference on Software Engineering: Software Engineering in Practice},
  pages={408--417},
  year={2024}
}

@inproceedings{jiang2024peek,
  title={A peek into token bias: Large language models are not yet genuine reasoners},
  author={Jiang, Bowen and Xie, Yangxinyu and Hao, Zhuoqun and Wang, Xiaomeng and Mallick, Tanwi and Su, Weijie J and Taylor, Camillo Jose and Roth, Dan},
  booktitle={Proceedings of the 2024 Conference on Empirical Methods in Natural Language Processing},
  pages={4722--4756},
  year={2024}
}

@inproceedings{shi2023large,
  title={Large language models can be easily distracted by irrelevant context},
  author={Shi, Freda and Chen, Xinyun and Misra, Kanishka and Scales, Nathan and Dohan, David and Chi, Ed H and Sch{\"a}rli, Nathanael and Zhou, Denny},
  booktitle={International Conference on Machine Learning},
  pages={31210--31227},
  year={2023},
  organization={PMLR}
}

@article{ouyang2022training,
  title={Training language models to follow instructions with human feedback},
  author={Ouyang, Long and Wu, Jeffrey and Jiang, Xu and Almeida, Diogo and Wainwright, Carroll and Mishkin, Pamela and Zhang, Chong and Agarwal, Sandhini and Slama, Katarina and Ray, Alex and others},
  journal={Advances in neural information processing systems},
  volume={35},
  pages={27730--27744},
  year={2022}
}

@article{bakhtin2019phyre,
  title={Phyre: A new benchmark for physical reasoning},
  author={Bakhtin, Anton and van der Maaten, Laurens and Johnson, Justin and Gustafson, Laura and Girshick, Ross},
  journal={Advances in Neural Information Processing Systems},
  volume={32},
  year={2019}
}

@inproceedings{wang2023newton,
  title={NEWTON: Are large language models capable of physical reasoning?},
  author={Wang, Yi and Duan, Jiafei and Fox, Dieter and Srinivasa, Siddhartha},
  booktitle={Findings of the association for computational linguistics: EMNLP 2023},
  pages={9743--9758},
  year={2023}
}

@article{cui2025curie,
  title={CURIE: Evaluating LLMs On Multitask Scientific Long Context Understanding and Reasoning},
  author={Cui, Hao and Shamsi, Zahra and Cheon, Gowoon and Ma, Xuejian and Li, Shutong and Tikhanovskaya, Maria and Norgaard, Peter and Mudur, Nayantara and Plomecka, Martyna and Raccuglia, Paul and others},
  journal={arXiv preprint arXiv:2503.13517},
  year={2025}
}

@article{chen2021codex,
  title={Evaluating Large Language Models Trained on Code},
  author={Mark Chen and Jerry Tworek and Heewoo Jun and Qiming Yuan and Henrique Ponde de Oliveira Pinto and Jared Kaplan and Harri Edwards and Yuri Burda and Nicholas Joseph and Greg Brockman and Alex Ray and Raul Puri and Gretchen Krueger and Michael Petrov and Heidy Khlaaf and Girish Sastry and Pamela Mishkin and Brooke Chan and Scott Gray and Nick Ryder and Mikhail Pavlov and Alethea Power and Lukasz Kaiser and Mohammad Bavarian and Clemens Winter and Philippe Tillet and Felipe Petroski Such and Dave Cummings and Matthias Plappert and Fotios Chantzis and Elizabeth Barnes and Ariel Herbert-Voss and William Hebgen Guss and Alex Nichol and Alex Paino and Nikolas Tezak and Jie Tang and Igor Babuschkin and Suchir Balaji and Shantanu Jain and William Saunders and Christopher Hesse and Andrew N. Carr and Jan Leike and Josh Achiam and Vedant Misra and Evan Morikawa and Alec Radford and Matthew Knight and Miles Brundage and Mira Murati and Katie Mayer and Peter Welinder and Bob McGrew and Dario Amodei and Sam McCandlish and Ilya Sutskever and Wojciech Zaremba},
  year={2021},
  eprint={2107.03374},
  archivePrefix={arXiv},
  primaryClass={cs.LG}
}

@article{austin2021program,
  title={Program synthesis with large language models},
  author={Austin, Jacob and Odena, Augustus and Nye, Maxwell and Bosma, Maarten and Michalewski, Henryk and Dohan, David and Jiang, Ellen and Cai, Carrie and Terry, Michael and Le, Quoc and others},
  journal={arXiv preprint arXiv:2108.07732},
  year={2021}
}

@article{jimenez2023swe,
  title={Swe-bench: Can language models resolve real-world github issues?},
  author={Jimenez, Carlos E and Yang, John and Wettig, Alexander and Yao, Shunyu and Pei, Kexin and Press, Ofir and Narasimhan, Karthik},
  journal={arXiv preprint arXiv:2310.06770},
  year={2023}
}

@article{glazer2024frontiermath,
  title={Frontiermath: A benchmark for evaluating advanced mathematical reasoning in ai},
  author={Glazer, Elliot and Erdil, Ege and Besiroglu, Tamay and Chicharro, Diego and Chen, Evan and Gunning, Alex and Olsson, Caroline Falkman and Denain, Jean-Stanislas and Ho, Anson and Santos, Emily de Oliveira and others},
  journal={arXiv preprint arXiv:2411.04872},
  year={2024}
}

@article{gottweis2025towards,
  title={Towards an AI co-scientist},
  author={Gottweis, Juraj and Weng, Wei-Hung and Daryin, Alexander and Tu, Tao and Palepu, Anil and Sirkovic, Petar and Myaskovsky, Artiom and Weissenberger, Felix and Rong, Keran and Tanno, Ryutaro and others},
  journal={arXiv preprint arXiv:2502.18864},
  year={2025}
}

@article{liu2023pre,
  title={Pre-train, prompt, and predict: A systematic survey of prompting methods in natural language processing},
  author={Liu, Pengfei and Yuan, Weizhe and Fu, Jinlan and Jiang, Zhengbao and Hayashi, Hiroaki and Neubig, Graham},
  journal={ACM computing surveys},
  volume={55},
  number={9},
  pages={1--35},
  year={2023},
  publisher={ACM New York, NY}
}

@article{tian2024scicode,
  title={Scicode: A research coding benchmark curated by scientists},
  author={Tian, Minyang and Gao, Luyu and Zhang, Shizhuo and Chen, Xinan and Fan, Cunwei and Guo, Xuefei and Haas, Roland and Ji, Pan and Krongchon, Kittithat and Li, Yao and others},
  journal={Advances in Neural Information Processing Systems},
  volume={37},
  pages={30624--30650},
  year={2024}
}

@article{jiang2025deepseek,
  title={DeepSeek vs. ChatGPT vs. Claude: A comparative study for scientific computing and scientific machine learning tasks},
  author={Jiang, Qile and Gao, Zhiwei and Karniadakis, George Em},
  journal={Theoretical and Applied Mechanics Letters},
  volume={15},
  number={3},
  pages={100583},
  year={2025},
  publisher={Elsevier}
}

@article{huang2020dynamic,
  title={Dynamic simulation of articulated soft robots},
  author={Huang, Weicheng and Huang, Xiaonan and Majidi, Carmel and Jawed, M Khalid},
  journal={Nature communications},
  volume={11},
  number={1},
  pages={2233},
  year={2020},
  publisher={Nature Publishing Group UK London}
}

@article{lejeune2020mechanical,
  title={Mechanical MNIST: A benchmark dataset for mechanical metamodels},
  author={Lejeune, Emma},
  journal={Extreme Mechanics Letters},
  volume={36},
  pages={100659},
  year={2020},
  publisher={Elsevier}
}

@article{niederer2021scaling,
  title={Scaling digital twins from the artisanal to the industrial},
  author={Niederer, Steven A and Sacks, Michael S and Girolami, Mark and Willcox, Karen},
  journal={Nature Computational Science},
  volume={1},
  number={5},
  pages={313--320},
  year={2021},
  publisher={Nature Publishing Group US New York}
}

@article{talischi2012polytop,
  title={PolyTop: a Matlab implementation of a general topology optimization framework using unstructured polygonal finite element meshes},
  author={Talischi, Cameron and Paulino, Glaucio H and Pereira, Anderson and Menezes, Ivan FM},
  journal={Structural and Multidisciplinary Optimization},
  volume={45},
  number={3},
  pages={329--357},
  year={2012},
  publisher={Springer}
}

@article{danabasoglu2020community,
  title={The community earth system model version 2 (CESM2)},
  author={Danabasoglu, Gokhan and Lamarque, J-F and Bacmeister, J and Bailey, DA and DuVivier, AK and Edwards, Jim and Emmons, LK and Fasullo, John and Garcia, R and Gettelman, Andrew and others},
  journal={Journal of Advances in Modeling Earth Systems},
  volume={12},
  number={2},
  pages={e2019MS001916},
  year={2020},
  publisher={Wiley Online Library}
}

@article{peng2011reproducible,
  title={Reproducible research in computational science},
  author={Peng, Roger D},
  journal={Science},
  volume={334},
  number={6060},
  pages={1226--1227},
  year={2011},
  publisher={American Association for the Advancement of Science}
}

@book{oberkampf2010verification,
  title={Verification and validation in scientific computing},
  author={Oberkampf, William L and Roy, Christopher J},
  year={2010},
  publisher={Cambridge university press}
}

@article{arndt2023deal,
  title={The deal. II library, version 9.5},
  author={Arndt, Daniel and Bangerth, Wolfgang and Bergbauer, Maximilian and Feder, Marco and Fehling, Marc and Heinz, Johannes and Heister, Timo and Heltai, Luca and Kronbichler, Martin and Maier, Matthias and others},
  journal={Journal of Numerical Mathematics},
  volume={31},
  number={3},
  pages={231--246},
  year={2023},
  publisher={De Gruyter}
}

@article{alnaes2015fenics,
  title={The FEniCS project version 1.5},
  author={Aln{\ae}s, Martin and Blechta, Jan and Hake, Johan and Johansson, August and Kehlet, Benjamin and Logg, Anders and Richardson, Chris and Ring, Johannes and Rognes, Marie E and Wells, Garth N},
  journal={Archive of numerical software},
  volume={3},
  number={100},
  year={2015}
}

@book{hughes2003finite,
  title={The finite element method: linear static and dynamic finite element analysis},
  author={Hughes, Thomas JR},
  year={2003},
  publisher={Courier Corporation}
}

@article{zienkiewicz1997finite,
  title={The finite element patch test revisited a computer test for convergence, validation and error estimates},
  author={Zienkiewicz, OC and Taylor, Richard Lawrence},
  journal={Computer methods in applied mechanics and engineering},
  volume={149},
  number={1-4},
  pages={223--254},
  year={1997},
  publisher={Elsevier}
}

@article{courant1994variational,
  title={Variational methods for the solution of problems of equilibrium and vibrations},
  author={Courant, Richard and others},
  journal={Lecture notes in pure and applied mathematics},
  pages={1--1},
  year={1994},
  publisher={MARCEL DEKKER AG}
}

@article{shojaei2025ai,
  title={AI-University: An LLM-based platform for instructional alignment to scientific classrooms},
  author={Shojaei, Mostafa Faghih and Gulati, Rahul and Jasperson, Benjamin A and Wang, Shangshang and Cimolato, Simone and Cao, Dangli and Neiswanger, Willie and Garikipati, Krishna},
  journal={arXiv preprint arXiv:2504.08846},
  year={2025}
}

@misc{garikipati_fem_coursera,
  author       = {Garikipati, Krishna},
  title        = {The Finite Element Method for Problems in Physics},
  howpublished = {Coursera [MOOC]},
  organization = {University of Michigan},
  year         = {2015},
  url          = {https://www.coursera.org/learn/finite-element-method}
}

@article{kamarei2026nine,
  title={Nine circles of elastic brittle fracture: A series of challenge problems to assess fracture models},
  author={Kamarei, Farhad and Zeng, Bo and Dolbow, John E and Lopez-Pamies, Oscar},
  journal={Computer Methods in Applied Mechanics and Engineering},
  volume={448},
  pages={118449},
  year={2026},
  publisher={Elsevier}
}

@article{szabo2021finite,
  title={Finite element analysis: Method, verification and validation},
  author={Szab{\'o}, Barna and Babu{\v{s}}ka, Ivo},
  year={2021},
  publisher={John Wiley \& Sons}
}

@book{gurtin2010mechanics,
  title={The mechanics and thermodynamics of continua},
  author={Gurtin, Morton E and Fried, Eliot and Anand, Lallit},
  year={2010},
  publisher={Cambridge university press}
}

@book{belytschko2014nonlinear,
  title={Nonlinear finite elements for continua and structures},
  author={Belytschko, Ted and Liu, Wing Kam and Moran, Brian and Elkhodary, Khalil},
  year={2014},
  publisher={John wiley \& sons}
}

@book{langtangen2003computational,
  title={Computational partial differential equations: numerical methods and diffpack programming},
  author={Langtangen, Hans Petter},
  volume={2},
  year={2003},
  publisher={Springer Berlin}
}

@book{mcguire2000matrix,
  title={Matrix structural analysis},
  author={McGuire, William and Gallagher, Richard H and Ziemian, Ronald D},
  year={2000}
}

@book{argyris1960energy,
  title={Energy theorems and structural analysis},
  author={Argyris, John H and Kelsey, Sydney and others},
  volume={60},
  year={1960},
  publisher={Springer}
}

@article{turner1956stiffness,
  title={Stiffness and deflection analysis of complex structures},
  author={Turner, M Jon and Clough, Ray W and Martin, Harold C and Topp, LJ},
  journal={journal of the Aeronautical Sciences},
  volume={23},
  number={9},
  pages={805--823},
  year={1956}
}

@book{cook2007concepts,
  title={Concepts and applications of finite element analysis},
  author={Cook, Robert D and others},
  year={2007},
  publisher={John wiley \& sons}
}

@article{felippa2004introduction,
  title={Introduction to finite element methods},
  author={Felippa, Carlos A},
  year={2004}
}

@book{strang1973analysis,
  title={An analysis of the finite element method},
  author={Strang, Gilbert and Fix, George J and others},
  volume={212},
  year={1973},
  publisher={Prentice-hall}
}

@book{press2007numerical,
  title={Numerical recipes 3rd edition: The art of scientific computing},
  author={Press, William H},
  year={2007},
  publisher={Cambridge university press}
}

@misc{bathe1996finite,
  title={Finite element procedures},
  author={Bathe, Klaus Jurgen},
  year={1996},
  publisher={Prentice Hall New Jersey}
}

@book{cottrell2009isogeometric,
  title={Isogeometric analysis: toward integration of CAD and FEA},
  author={Cottrell, J Austin and Hughes, Thomas JR and Bazilevs, Yuri},
  year={2009},
  publisher={John Wiley \& Sons}
}

@book{higham2002accuracy,
  title={Accuracy and stability of numerical algorithms},
  author={Higham, Nicholas J},
  year={2002},
  publisher={SIAM}
}

@misc{holzapfel2002nonlinear,
  title={Nonlinear solid mechanics: a continuum approach for engineering science},
  author={Holzapfel, Gerhard A},
  year={2002},
  publisher={Kluwer Academic Publishers Dordrecht}
}

@article{macneal1985proposed,
  title={A proposed standard set of problems to test finite element accuracy},
  author={Macneal, Richard H and Harder, Robert L},
  journal={Finite elements in analysis and design},
  volume={1},
  number={1},
  pages={3--20},
  year={1985},
  publisher={Elsevier}
}

@book{roache1998verification,
  title={Verification and validation in computational science and engineering},
  author={Roache, Patrick J},
  volume={895},
  year={1998},
  publisher={Hermosa Albuquerque, NM}
}

@book{ammann2017introduction,
  title={Introduction to software testing},
  author={Ammann, Paul and Offutt, Jeff},
  year={2017},
  publisher={Cambridge University Press}
}

@book{foley1996computer,
  title={Computer graphics: principles and practice},
  author={Foley, James D},
  volume={12110},
  year={1996},
  publisher={Addison-Wesley Professional}
}

@book{trefethen2022numerical,
  title={Numerical linear algebra},
  author={Trefethen, Lloyd N and Bau, David},
  year={2022},
  publisher={SIAM}
}

@article{hamdi2026towards,
  title={Towards robust surrogate models: Benchmarking machine learning approaches to expediting phase field simulations of brittle fracture},
  author={Hamdi, Erfan and Lejeune, Emma},
  journal={Computer Methods in Applied Mechanics and Engineering},
  volume={449},
  pages={118526},
  year={2026},
  publisher={Elsevier}
}

@article{mudur2024feabench,
  title={Feabench: Evaluating language models on real world physics reasoning ability},
  author={Mudur, Nayantara and Cui, Hao and Venugopalan, Subhashini and Raccuglia, Paul and Brenner, Michael and Norgaard, Peter Christian},
  year={2024}
}

@article{song2025evaluating,
  title={Evaluating Large Language Models in Scientific Discovery},
  author={Song, Zhangde and Lu, Jieyu and Du, Yuanqi and Yu, Botao and Pruyn, Thomas M and Huang, Yue and Guo, Kehan and Luo, Xiuzhe and Qu, Yuanhao and Qu, Yi and others},
  journal={arXiv preprint arXiv:2512.15567},
  year={2025}
}

@article{shojaee2025llm,
  title={Llm-srbench: A new benchmark for scientific equation discovery with large language models},
  author={Shojaee, Parshin and Nguyen, Ngoc-Hieu and Meidani, Kazem and Farimani, Amir Barati and Doan, Khoa D and Reddy, Chandan K},
  journal={arXiv preprint arXiv:2504.10415},
  year={2025}
}

@inproceedings{cai2025sciassess,
  title={Sciassess: Benchmarking llm proficiency in scientific literature analysis},
  author={Cai, Hengxing and Cai, Xiaochen and Chang, Junhan and Li, Sihang and Yao, Lin and Changxin, Wang and Gao, Zhifeng and Wang, Hongshuai and Yongge, Li and Lin, Mujie and others},
  booktitle={Findings of the Association for Computational Linguistics: NAACL 2025},
  pages={2335--2357},
  year={2025}
}

@article{liang2022holistic,
  title={Holistic evaluation of language models},
  author={Liang, Percy and Bommasani, Rishi and Lee, Tony and Tsipras, Dimitris and Soylu, Dilara and Yasunaga, Michihiro and Zhang, Yian and Narayanan, Deepak and Wu, Yuhuai and Kumar, Ananya and others},
  journal={arXiv preprint arXiv:2211.09110},
  year={2022}
}

@book{zienkiewicz2005finite,
  title={The finite element method set},
  author={Zienkiewicz, Olgierd C and Taylor, Robert Leroy},
  year={2005},
  publisher={Elsevier}
}

@article{pandey2026beyond,
  title={Beyond the Answer: Decoding the Behavior of LLMs as Scientific Reasoners},
  author={Pandey, Rohan and Ye, Eric and Li, Michael},
  journal={arXiv preprint arXiv:2603.28038},
  year={2026}
}

@article{deotale2026all,
  title={All-fem: Agentic large language models fine-tuned for finite element methods},
  author={Deotale, Rushikesh and Srinivasan, Adithya and Golestanian, Mahmoud and Tian, Yuan and Zhang, Tianyi and Vlachos, Pavlos and Gomez, Hector},
  journal={Computer Methods in Applied Mechanics and Engineering},
  volume={457},
  pages={118985},
  year={2026},
  publisher={Elsevier}
}

@inproceedings{xia2025evaluating,
  title={Evaluating mathematical reasoning beyond accuracy},
  author={Xia, Shijie and Li, Xuefeng and Liu, Yixin and Wu, Tongshuang and Liu, Pengfei},
  booktitle={Proceedings of the AAAI Conference on Artificial Intelligence},
  volume={39},
  number={26},
  pages={27723--27730},
  year={2025}
}

@article{hao2024llm,
  title={Llm reasoners: New evaluation, library, and analysis of step-by-step reasoning with large language models},
  author={Hao, Shibo and Gu, Yi and Luo, Haotian and Liu, Tianyang and Shao, Xiyan and Wang, Xinyuan and Xie, Shuhua and Ma, Haodi and Samavedhi, Adithya and Gao, Qiyue and others},
  journal={arXiv preprint arXiv:2404.05221},
  year={2024}
}

@inproceedings{tong2024codejudge,
  title={Codejudge: Evaluating code generation with large language models},
  author={Tong, Weixi and Zhang, Tianyi},
  booktitle={Proceedings of the 2024 Conference on Empirical Methods in Natural Language Processing},
  pages={20032--20051},
  year={2024}
}

@inproceedings{yang2025llm,
  title={How is llm reasoning distracted by irrelevant context? an analysis using a controlled benchmark},
  author={Yang, Minglai and Huang, Ethan and Zhang, Liang and Surdeanu, Mihai and Wang, William Yang and Pan, Liangming},
  booktitle={Proceedings of the 2025 Conference on Empirical Methods in Natural Language Processing},
  pages={13340--13358},
  year={2025}
}

@article{hang2026pdeagent,
  title={PDEAgent-Bench: A Multi-Metric, Multi-Library Benchmark for PDE Solver Generation},
  author={Hang, Zhen and Yashengjiang, Yushan and Li, Junhui and Dong, Huanshuo and Wei, Yang and Hao, Zhezheng and Ma, Jiangtao and Bai, Songlin and Kai, Haozhong and Yue, Xihang and others},
  journal={arXiv preprint arXiv:2605.09636},
  year={2026}
}

\end{document}